\pdfoutput=1

\documentclass[11pt]{article}

\usepackage[final]{acl}

\usepackage{algorithmic}
\usepackage{algorithm}
\usepackage{subfigure}

\usepackage{times}
\usepackage{latexsym}
\usepackage{multirow}
\usepackage{graphicx}
\usepackage{booktabs}
\usepackage{amsmath}
\usepackage{amsfonts}
\usepackage{tabularx}
\usepackage{bm}
\definecolor{greenberry}{HTML}{0aa344}
\usepackage[T1]{fontenc}

\usepackage[utf8]{inputenc}

\usepackage{microtype}

\usepackage{inconsolata}

\title{Strengthened Symbol Binding Makes Large Language Models Reliable Multiple-Choice Selectors}

\author{Mengge Xue\footnotemark[1],\quad Zhenyu Hu\footnotemark[1],\quad Liqun Liu\footnotemark[2], \\ {\bf Kuo Liao,} \quad {\bf Shuang Li,}\quad {\bf Honglin Han,}\quad {\bf Meng Zhao,}\quad {\bf Chengguo Yin} \\
Tencent\\
\texttt{\{berryxue, mapleshu, liqunliu\}@tencent.com}}

\begin{document}
\maketitle

\begin{abstract}
\footnotetext[1]{These authors contributed equally to this work.}
\footnotetext[2]{Corresponding author.}

Multiple-Choice Questions (MCQs) constitute a critical area of research in the study of Large Language Models (LLMs).
Previous works have investigated the selection bias problem in MCQs within few-shot scenarios, in which the LLM's performance may be influenced by the presentation of answer choices, leaving the selection bias during Supervised Fine-Tuning (SFT) unexplored.
In this paper, we reveal that selection bias persists in the SFT phase , primarily due to the LLM's inadequate Multiple Choice Symbol Binding (MCSB) ability. 
This limitation implies that the model struggles to associate the answer options with their corresponding symbols (e.g., A/B/C/D) effectively.
To enhance the model's MCSB capability, 
we first incorporate option contents into the loss function and subsequently adjust the weights of the option symbols and contents, guiding the model to understand the option content of the current symbol.
Based on this, we introduce an efficient SFT algorithm for MCQs, termed Point-wise Intelligent Feedback (PIF). PIF constructs negative instances by randomly combining the incorrect option contents with all candidate symbols, and proposes a point-wise loss to provide feedback on these negative samples into LLMs. 
Our experimental results demonstrate that PIF significantly reduces the model's selection bias by improving its MCSB capability. Remarkably, PIF exhibits a substantial enhancement in the accuracy for MCQs\footnote[3]{The code of this work is available at \url{https://github.com/berryxue/PIF}}.

\end{abstract}

\section{Introduction}

\begin{figure}[t]
\centering
\includegraphics[width=0.48\textwidth]{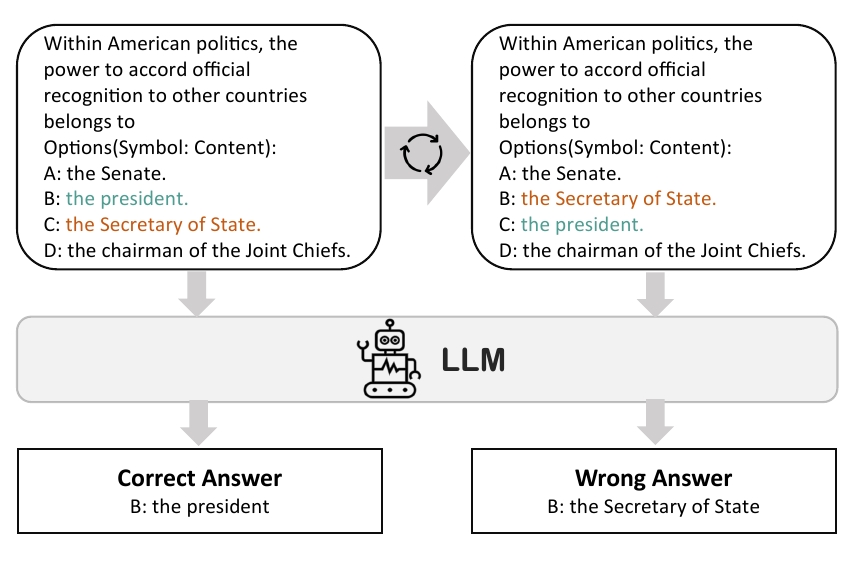}
\caption{Selection bias of MCQs. Upon transposition of the correct content from option B to C, the model persists in selecting B instead of the correct option content.}
\label{fig:introduction}
\end{figure}

Multiple-Choice Questions (MCQs) are ubiquitously employed in the realm of Large Language models (LLMs). They typical comprise a query accompanied by an array of potential options, wherein the model's assignment is to discern and select the most appropriate solution. 
Given the significance, numerous studies \citep{pezeshkpour2023large,zheng2023large,robinson2023iclr} have explored the challenges faced in the few-shot phase of MCQs. One of the main challenges is the selection bias \citep{pezeshkpour2023large,zheng2023large}, which refers that LLMs are sensitive to variations in the arrangement of options within MCQs. Figure~\ref{fig:introduction} demonstrates an example of selection bias.

Foundation LLMs such as LLaMA2-7B can seldom acquire data-sensitive domain knowledge.
For such scenarios, it is crucial to perform Supervised Fine-Tuning (SFT) on LLMs.
Thus we hope LLMs to ensure accuracy and also robustly select reliable options in MCQs during SFT. 
Unfortunately, we find that the selection bias still exists during the SFT phase. Following the methodology of \citet{zheng2023large}, we conduct “\textit{answer-moving attack}” experiment by always moving the golden option content to a specific symbol, and the results are displayed in Table~\ref{table:ablation_feature}.
In detail, we train two models using the training dataset, and select the best-performing model on the validation set and predict results for the test set. Then, we implement the answer-moving attack by moving all the correct options to the same symbol for the test dataset and predict results again, which cause significant changes in the models' performance. For example, when moving all correct options to symbol A, the accuracy of the LLaMA2-7B model increases by 11.1, while LLaMA2-13B model decreases 4.4.

\begin{table}
\centering
\scalebox{0.75}{
  \begin{tabular}{llllllll}
    \toprule
    \multicolumn{1}{c}{\textbf{dataset}} & \multicolumn{5}{c}{\textbf{MMLU}}  \\
\cmidrule(lr){2-6}

\multicolumn{1}{c}{\textbf{Move Golden to}} & \multicolumn{1}{c}{\textbf{Orig}} & \multicolumn{1}{c}{\textbf{A}} & \multicolumn{1}{c}{\textbf{B}} & \multicolumn{1}{c}{\textbf{C}} & \multicolumn{1}{c}{\textbf{D}}  \\ \hline
  \multicolumn{1}{c}{\multirow{2}*{LLaMA2-7B}} & \multicolumn{1}{c}{\multirow{2}*{54.6}} & \multicolumn{1}{c}{65.7} & \multicolumn{1}{c}{45.6} & \multicolumn{1}{c}{58.4} & \multicolumn{1}{c}{47.8}  \\
 ~&~& \multicolumn{1}{c}{\textcolor{red}{(+11.1)}} & \multicolumn{1}{c}{\textcolor{greenberry}{(-9.0)}} & \multicolumn{1}{c}{\textcolor{red}{(+3.8)}} & \multicolumn{1}{c}{\textcolor{greenberry}{(-6.8)}}  \\ \hline
  \multicolumn{1}{c}{\multirow{2}*{LLaMA2-13B}} & \multicolumn{1}{c}{\multirow{2}*{59.2}} & \multicolumn{1}{c}{54.8} & \multicolumn{1}{c}{64.6} & \multicolumn{1}{c}{56.3} & \multicolumn{1}{c}{61.5}  \\
 ~&~& \multicolumn{1}{c}{\textcolor{greenberry}{(-4.4)}} & \multicolumn{1}{c}{\textcolor{red}{(+5.4)}} & \multicolumn{1}{c}{\textcolor{greenberry}{(-2.9)}} & \multicolumn{1}{c}{\textcolor{red}{(+2.3)}} \\  
    \bottomrule
  \end{tabular}
}
\caption{\label{table:ablation_feature}
    The accuracy results after answer-moving attack on the LLMs during the SFT process with MMLU benchmarks. Relocating the correct option content of MCQs to a specific symbol can lead to significant performance fluctuations for LLMs.
}
\end{table}

Why do LLMs show selection bias during the SFT stage? 
We propose a hypothesis that \textbf{\textit{"Strengthened Symbol Binding Makes Large Language Models Reliable Multiple-Choice Selectors"}}.
We utilize \textbf{M}ultiple \textbf{C}hoice \textbf{S}ymbol \textbf{B}inding (MCSB) capability~\citep{robinson2023iclr} to represent the LLM’s ability to bind option symbols and their corresponding option contents, and employ \textbf{P}roportion of \textbf{P}lurality \textbf{A}greement (PPA) metric to compare the relative MCSB ability of two LLMs. 
Through comprehensive experimental validation, we discover that 
improving the LLMs' performance on the PPA metric alleviates the LLMs' performance on selection bias.


Based on the relationship between the LLMs’ selection bias and its MCSB capabilities, we expect to mitigate selection bias by enhancing the model's MCSB capability. 
We first incorporate option contents into the loss function, guiding the model to understand the content of the current symbol. 
However, the results are less than satisfactory.
Considering that label words are anchors~\citep{wang-etal-2023-label},
we adjust the weights of the option symbols and contents in the optimization objective, termed Reweighting Symbol-Content Binding (RSCB).
Subsequently, inspired by Reinforcement Learning from Human Feedback~\citep{stiennon2020learning},  we propose Point-wise Intelligent Feedback (PIF), in which we construct negative samples by randomly combining the contents of incorrect options with all option symbols,
 and design a point-wise loss to feedback these negative samples into SFT.
Finally, PIF not only ensures the stability of model performance but also enhances it.

In summary, our contributions are as follows: 
(1) 
We conduct experiments to demonstrate that there is still the selection bias when performing SFT on LLMs for MCQs. 
We hypothesize that \textit{Strengthened Symbol Binding Makes Large Language Models Reliable Multiple-Choice Selectors}. 
In Section~\ref{sec:MCSB}, we validate this hypothesis by investigating the correlation between the MCSB capability and the selection bias. 
(2) We mitigate selection bias by enhancing the model’s MCSB capability. 
We propose the Point-wise Intelligent Feedback (PIF) method, which constructs negative samples by randomly combining the content of incorrect options with all candidate symbols and designing a point-wise loss to provide feedback on these negative samples into LLMs.
(3) We conduct extensive experiments to validate that our PIF method can significantly alleviate the selection bias of LLMs during the SFT phase for MCQs. Meanwhile, the experimental results prove that PIF could also enhance the accuracy performance of the model. 

\begin{figure*}[t]
\centering
\includegraphics[width=0.9\textwidth]{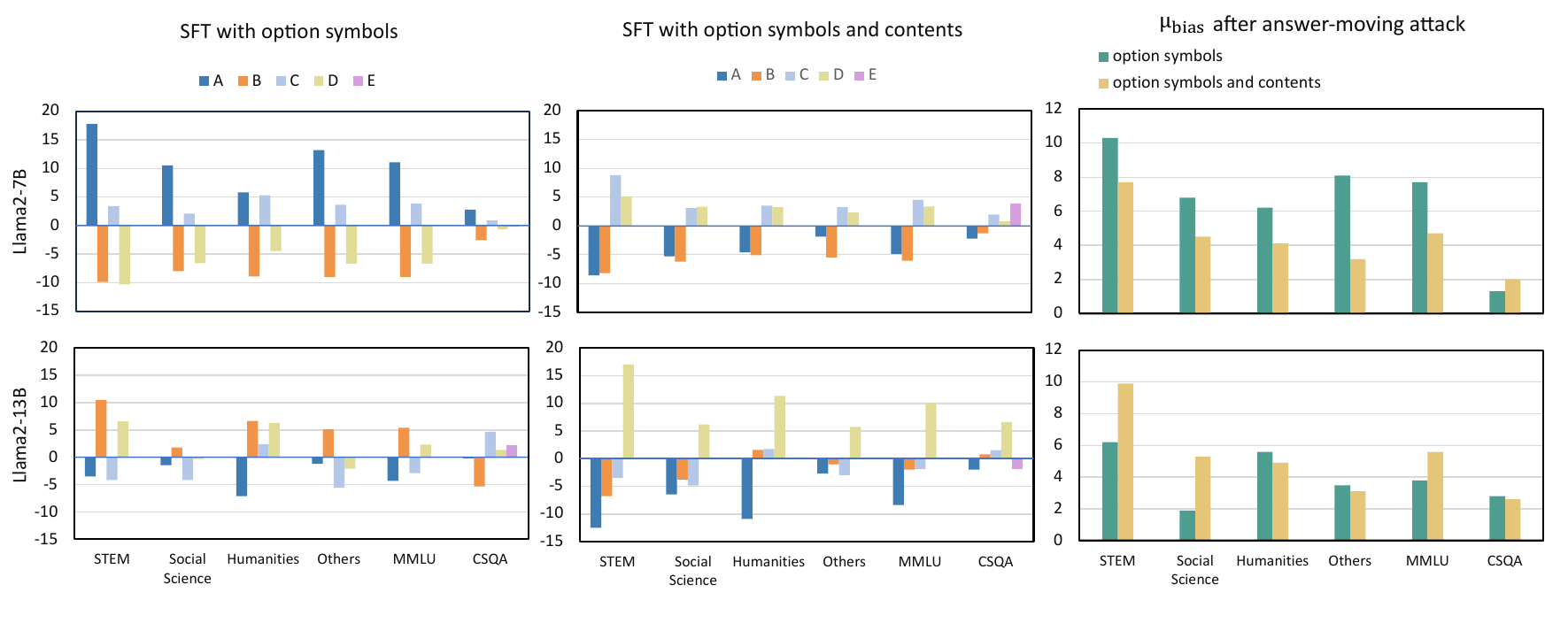}
\caption{LLMs' selection bias during SFT. The leftmost two columns demonstrate the changes in accuracy following the answer-moving attack, whereas the rightmost column exhibits the metric $\mu_\text{bias}$ as defined by Equation~\ref{eq:mubias}.}
\label{fig:basic bias}
\end{figure*}

\section{Exploration of Selection Bias During Supervised Fine-tuning}
\subsection{Experimental Background}
\noindent \textbf{Datasets}\quad
To investigate the selection bias in the SFT stage of LLMs for MCQs, we conduct experiments on several commonly used MCQs benchmarks: Massive Multitask Language Understanding(\textbf{MMLU}) benchmark \citep{hendryckstest2021} and CommonsenseQA(\textbf{CSQA}) \citep{talmor-etal-2019-commonsenseqa}. Our choice of benchmarks considers the variety of tasks and fields involved. Explicitly, MMLU comprises 4-option multiple-choice questions, whereas CSQA contains 5-option variants. As for domains, MMLU encompasses a wide range of tasks from fields including STEM, humanities, social science and others, coverage 57 subjects; while CSQA conduct questions draw upon ConceptNet \citep{speer2017} concepts, enabling variety and integration of worldly knowledge within the queries. It's worth noting that the CSQA dataset does not have standard answers for the test data. Thus in our experimental setup, we use the dev set as the test set and extract a portion of samples from the train set to serve as the validation set. MMLU is additionally split into four domains (STEM, Social Science, Humanities, Others) based on its subject categories. Details can be found in Appendix~\ref{sec:benchmark statistics}.

\noindent \textbf{Models}\quad
We conduct experiments on LLMs originating from well-known LLM families, spanning various sizes. \textbf{LLaMA} \citep{touvron2023llama} constitutes a compilation of widely-used foundation models designed to promote research on LLMs for training, inference, and extensive applications. In this paper, taking into account the model size, we assess the 7B and 13B variant of LLaMA2 (LLaMA2-7B, LLaMA2-13B). Projects used in this work is illustrated in Appendix~\ref{sec:llms project}.

Taking into account the limited computational resources, we fine-tune LLMs with \textbf{L}ow-\textbf{R}ank \textbf{A}daptation(LoRA) \citep{hu2022lora} and set its rank to 16, the alpha parameter to 64, and the dropout rate to 0.1. 
To investigate the selection bias, we design two distinct output configurations for LLMs during the training process. 
One generates only the option symbols, defined as $\pi_{\text{Symbol}}$, while the other produces both the option symbols and contents concurrently, defined as $\pi_{\text{SCB}}$.
During the testing process, the model is configured to output only the symbol, which is sufficient for determining whether the answer is correct or not.

\setlength{\tabcolsep}{4pt}
\begin{table*}
\begin{center}
\scalebox{0.78}{
\begin{tabular}{l|cc|cc|cc|cc|cc|cc}
\hline
\multirow{2}*{Model}&\multicolumn{2}{c|}{STEM}&\multicolumn{2}{c|}{Social Science} &\multicolumn{2}{c|}{Human} &\multicolumn{2}{c|}{Others}&\multicolumn{2}{c|}{MMLU}&\multicolumn{2}{c}{CSQA} \\ \cline{2-13}
~&$\Delta\mu_\text{bias}$&$\Delta\mu_\text{ppa}$&$\Delta\mu_\text{bias}$&$\Delta\mu_\text{ppa}$&$\Delta\mu_\text{bias}$&$\Delta\mu_\text{ppa}$&$\Delta\mu_\text{bias}$&$\Delta\mu_\text{ppa}$&$\Delta\mu_\text{bias}$&$\Delta\mu_\text{ppa}$&$\Delta\mu_\text{bias}$&$\Delta\mu_\text{ppa}$ \\  \hline \hline

LLaMA2-7B
&\textcolor{greenberry}{-2.6}&\textcolor{red}{+0.6}
&\textcolor{greenberry}{-2.3}&\textcolor{red}{+1.2}
&\textcolor{greenberry}{-2.1}&\textcolor{red}{+2.4}
&\textcolor{greenberry}{-4.9}&\textcolor{red}{+1.2}
&\textcolor{greenberry}{-3.0}&\textcolor{red}{+1.6}
&\textcolor{red}{+0.7}&\textcolor{greenberry}{-0.2} \\
\hline
LLaMA2-13B
&\textcolor{red}{+3.7}&\textcolor{greenberry}{-2.1}
&\textcolor{red}{+3.4}&\textcolor{greenberry}{-1.0}
&\textcolor{greenberry}{-0.7}&\textcolor{red}{+0.2}
&\textcolor{greenberry}{-0.4}&\textcolor{greenberry}{-0.3}
&\textcolor{red}{+1.8}&\textcolor{greenberry}{-0.7}
&\textcolor{greenberry}{-0.2}&\textcolor{red}{+0.3} \\
\hline
\end{tabular}
}
\end{center}
\caption{\label{table:bias ppa}Relationships between $\Delta\mu_\text{bias}$ and $\Delta\mu_\text{ppa}$. When $\Delta\mu_\text{bias}$ is positive, $\Delta\mu_\text{ppa}$ is negative and vice versa. Such indicating that "\textit{Strengthened Symbol Binding Makes Large Language Models Reliable Multiple-Choice Selectors}".}
\end{table*}
\setlength{\tabcolsep}{1.4pt}

\subsection{Selection Bias During SFT}
\label{sec:selection bias}
We conduct experiments using two LLMs on six datasets. 
Similar to \citet{zheng2023large}, we conduct \textit{"answer-moving attack"} experiment to measure the selection bias. During testing, we move all the correct answers to A|B|C|D|E respectively (Answers are A|B|C|D in MMLU. For the sake of convenience, we will use the unified notation of A|B|C|D|E) for the test set, and then display the model's accuracy after the relocation. Subsequently, to provide a concrete numerical representation of the impact of the answer-moving attack, 
we calculate $\mu_\text{bias}$, the average absolute value of the difference between the accuracy after the answer-moving attack and the original accuracy tested on the standard test set. Which can be formulated as:
\begin{equation}
\label{eq:mubias}
    \mu_\text{bias} = \frac{\sum_{i=1}^{K} (|\text{Acc}_i - \text{Acc}_0|)}{K},
\end{equation}
$\text{Acc}_i$ refers to the accuracy after answer-moving attack, $\text{Acc}_0$ is the accuracy on the standard testing set, $K$ is the number of options, and is set as 5 for CSQA, 4 for MMLU.
Through the experimental results in Figure~\ref{fig:basic bias}, we make following observations: 

\noindent \textbf{Selection bias varies among different sizes of the model parameters.}\quad As seen from Figure~\ref{fig:basic bias}, LLaMA2-7B tends to choose C, while LLaMA2-13B prefers to choose D. Although these two models belong to the same model family, their performances are not identical, which aligns with the findings in Paper \citet{zheng2023large}. We also observe that, when fine-tuning the model only with option symbols, LLaMA2-13B has a smaller $\mu_\text{bias}$ on MMLU compared to LLaMA2-7B, while reversed on the CSQA dataset. Therefore, during SFT, there is no absolute relationship between selection bias and the size of the model parameters.

\noindent \textbf{SFT training examples influence both the magnitude and distribution of selection bias.}\quad Regardless of the LLMs, the overall bias value $\mu_\text{bias}$ for CSQA is always smaller than that of MMLU. As seen from Appendix~\ref{sec:benchmark statistics}, the proportions of the five options in the CSQA benchmark training set are roughly equal, while this is not the case for MMLU. Additionally, when we fine-tune LLaMA2-13B with option symbols, the model shows a stronger inclination to predict option B on MMLU, while on CSQA, it predicts B with a significantly lower probability than others. Therefore, we speculate that the selection bias during SFT is closely related to the fine-tuning dataset used in the SFT stage.

\subsection{Why Do LLMs Suffer Selection Bias in MCQs' SFT}
\label{sec:MCSB}
After analyzing the selection bias during SFT in various LLMs and datasets, we now focus on understanding why LLMs exhibit selection bias during the SFT stage. 
We identify two reasons why LLMs may choose the wrong answer.
Firstly, the LLM does not know the correct option, which can be attributed to its capability. 
Secondly, the LLM is aware of the correct option content. However, selection bias makes it choose a preferred option symbol instead of one corresponding to the correct option. This indicates the model's weak ability to associate option content with the appropriate symbol. 
According to it, we propose a hypothesis:


\noindent\textbf{\textit{Strengthened Symbol Binding Makes Large Language Models Reliable Multiple-Choice Selectors.}}

In this paper, we utilize \textit{\textbf{M}ultiple \textbf{C}hoice \textbf{S}ymbol \textbf{B}inding }(MCSB) capability \citep{robinson2023iclr} to represent the LLM's ability to bind option symbols and their corresponding contents. 
Similar to \citet{robinson2023iclr}, we also employ the \textit{\textbf{P}roportion of \textbf{P}lurality \textbf{A}greement }($\mu_\text{ppa}$) metric to 
compare the relative MCSB ability of two LLMs.
\begin{equation}
\label{eq:mu_ppa}
    \mu_\text{ppa} = \frac{1}{|\mathbb{D}|}\sum\limits_{\mathbb{D}}\frac{\max\limits_{k\sim K}(\sum\limits_{j=1}\limits^{K!}y_j=o_k)}{K!}.
\end{equation}

As shown in Table~\ref{table:ppa cal}, given a question with $K$ options, there are $K!$ distinct arrangements of these options in a fixed set of symbols. 
For ease of expression, we use $o_k$ to represent the $k$-th option content and $y_j$ to describe the content of the predicted arrangement.
During testing, we present each question to the model using each unique ordering, and then PPA for this question is the frequency corresponding to the most frequently predicted option content.
For a dataset $\mathbb{D}$, $\mu_\text{ppa}$ is calculated as the average of the PPAs for all individual questions. 

To demonstrate the relationship between the MCSB capability and the selection bias, we calculate the differences of $\mu_\text{ppa}$ and $\mu_\text{bias}$ between $\pi_\text{Symbol}$ and $\pi_\text{SCB}$, which are defined as follows:

\begin{equation}
\label{eq:Deltamu}
    \Delta\mu_\text{bias} = \mu_\text{bias}^{\pi_\text{SCB}}-\mu_\text{bias}^{\pi_\text{Symbol}},
\end{equation}
$\Delta\mu_\text{ppa}$ is defined in a similar approach. As delineated in the prior analysis, if there exists a correlation between selection bias and the LLM's MCSB capacity, 
then a positive $\Delta\mu_\text{ppa}$ corresponds to a negative $\Delta\mu_\text{bias}$, indicating that increasing the MCSB capability resulting a reduction in the LLM's bias. 
We conduct experiments with LLaMA2-7B and LLaMA2-13B. The result is illustrated in Table~\ref{table:bias ppa}.
Except for the performance of LLaMA2-13B on MMLU-Others, all the remaining results indicate that when $\Delta\mu_\text{bias}$ is positive, $\Delta\mu_\text{ppa}$ is negative and vice versa. Additionally, the performance of LLaMA2-13B on MMLU-Others can be considered as having a relatively constant $\mu_\text{ppa}$, rather than being contrary to our conclusion.
Appendix~\ref{sec:relationship proof} also demonstrates a theoretical proof.

\section{Methodology}
\label{sec:method}
According to the previous analysis, we can mitigate the selection bias of LLMs during SFT by enhancing their MCSB abilities.

\begin{figure*}[t]
\centering
\includegraphics[width=0.9\textwidth]{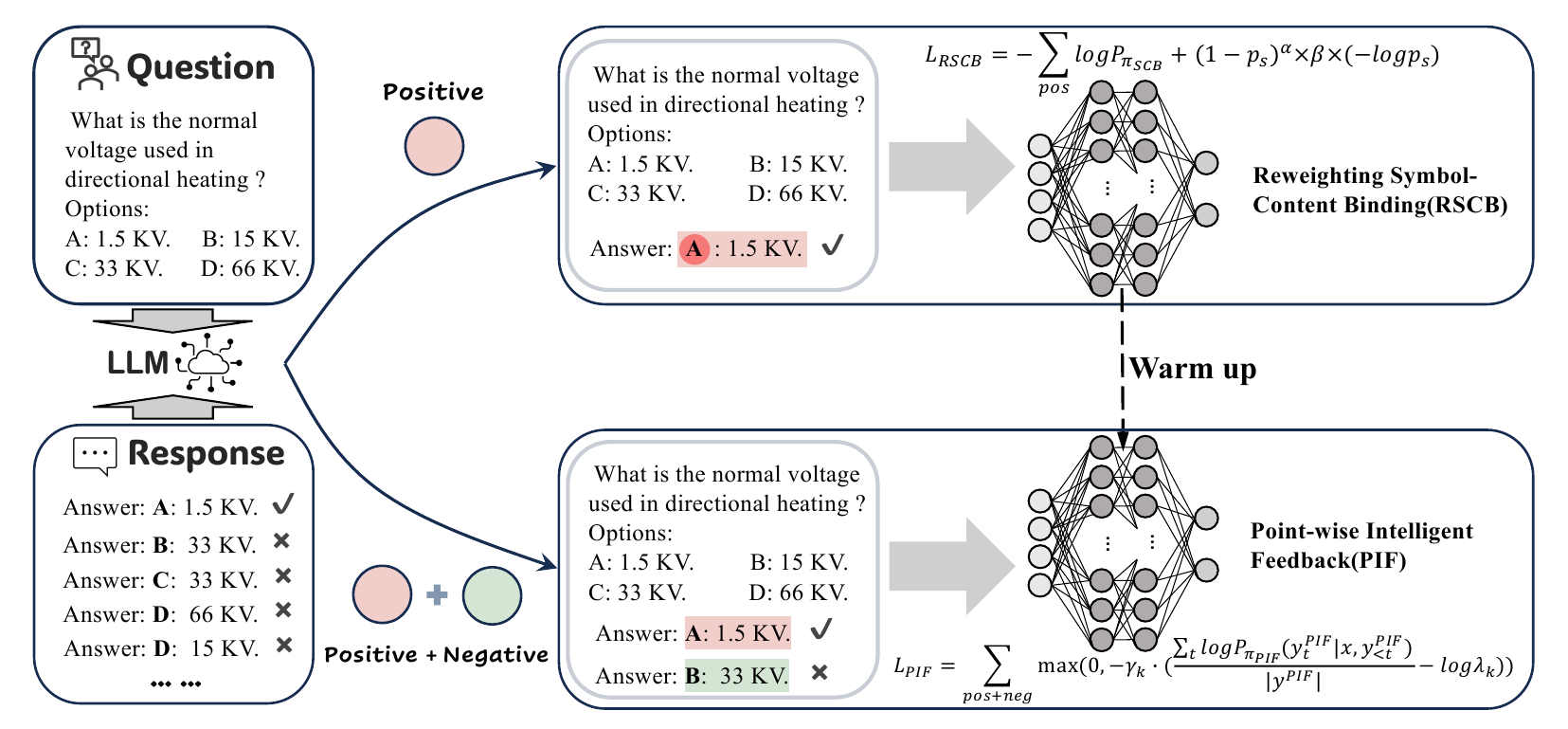}
\caption{Visualization of RSCB and PIF. RSCB adjusts the weights of the option symbols and contents in the SFT optimization objective. PIF constructs negative samples by randomly combining the content of incorrect options with all option symbols and designs a point-wise loss to feedback these negative samples into SFT.}
\label{fig:model backbone}
\end{figure*}

\subsection{Symbol-Content Binding}

Initially, we incorporate the option symbols as the LLM's target tokens during training, resulting in a model $\pi_\text{Symbol}$. Given input $x$ and output $y^\text{Symbol}$, we define optimization objective as:

\begin{equation}
\label{eq:s_sft}
\mathcal{L}_{\text{Symbol}} = -\frac{\sum_t {\log} P_{\pi_{\text{Symbol}}}(y^\text{Symbol}_t|x,y^\text{Symbol}_{<t})}{|y^\text{Symbol}|}.
\end{equation}

As shown in Table~\ref{table:ablation_feature}, $\pi_\text{Symbol}$ suffers from severe selection bias. Since enhancing the MCSB capabilities can effectively alleviate the selection bias, given output $y^\text{SCB}$, we propose the \textbf{S}ymbol-\textbf{C}ontent \textbf{B}inding (SCB) debiasing method $\pi_{\text{SCB}}$, 
which incorporates both the option symbols and contents as the LLM's target tokens during training:

\begin{equation}
\label{eq:sc_sft}
\mathcal{L}_{\text{SCB}} = -\frac{\sum_t {\log} P_{\pi_\text{SCB}}(y^\text{SCB}_t|x,y^\text{SCB}_{<t})}{|y^\text{SCB}|}.
\end{equation}

However, the results are not as expected. As shown in Table~\ref{table:bias ppa}, there is no clear pattern indicating that $\pi_{\text{SCB}}$ has a lower bias compared to $\pi_{\text{Symbol}}$.

\subsection{Reweighting Symbol-Content Binding}
Actually, label words are anchors~\citep{wang-etal-2023-label}, which LLMs pay more attention to. On the other hand, the answer content merely plays an auxiliary role, assisting the model in comprehending the actual content of the corresponding symbol. Thus we adjust the weights of the option symbols and contents in the optimization objective, termed \textbf{R}eweighting \textbf{S}ymbol-\textbf{C}ontent \textbf{B}inding (RSCB). The objective function is defined as:
\begin{equation}
\label{eq:bsc_sft}
\mathcal{L}_{\text{RSCB}} = \mathcal{L}_{\text{SCB}}+(1-p_\text{s})^\alpha\cdot\beta\cdot(-\log p_\text{s}).
\end{equation}
Where $p_\text{s}$ is the predicted probability of the symbol token, $\beta$  is the re-assigned weight for the symbol token. Considering that LLM itself has already focused on the symbol tokens for simple samples, there is no need to emphasize the symbol tokens specifically in such cases. Therefore, we ultimately employ the Focal loss~\citep{lin2017focal}.

\setlength{\tabcolsep}{4pt}
\begin{table*}
\begin{center}
\scalebox{0.83}{
\begin{tabular}{llcccccccccccc}
\toprule\noalign{\smallskip}

\multirow{2}*{\textbf{Model}}&\multirow{2}*{\textbf{Method}}&\multicolumn{2}{c}{\textbf{STEM}}&\multicolumn{2}{c}{\textbf{Social Science}}&\multicolumn{2}{c}{\textbf{Humanities}}&\multicolumn{2}{c}{\textbf{Others}}&\multicolumn{2}{c}{\textbf{MMLU}}&\multicolumn{2}{c}{\textbf{CSQA}} \\ \cmidrule(lr){3-4} \cmidrule(lr){5-6} \cmidrule(lr){7-8} \cmidrule(lr){9-10} \cmidrule(lr){11-12} \cmidrule(lr){13-14} 

~&~&$\mu_\text{bias}$$\downarrow$&$\mu_\text{ppa}$$\uparrow$&$\mu_\text{bias}$$\downarrow$&$\mu_\text{ppa}$$\uparrow$&$\mu_\text{bias}$$\downarrow$&$\mu_\text{ppa}$$\uparrow$&$\mu_\text{bias}$$\downarrow$&$\mu_\text{ppa}$$\uparrow$&$\mu_\text{bias}$$\downarrow$&$\mu_\text{ppa}$$\uparrow$&$\mu_\text{bias}$$\downarrow$&$\mu_\text{ppa}$$\uparrow$  \\ 
\noalign{\smallskip} \hline

\multirow{5}*{LLaMA2-7B}&Symbol&10.3&70.9&6.8&81.8&6.2&81.3&8.1&80.2&7.7&78.9&1.3&93.1\\ \cline{2-14}

~&\multirow{1}*{SCB}
&7.7&71.5&4.5&83&4.1&83.7&3.2&81.4&4.7&80.5&2.0&92.9 \\

~&\multirow{1}*{RSCB}&3.8&71.8&3.4&83.1&5.3&81.8&2.9&81.6&3.0&79.8&2.4&92.4\\

~&\multirow{2}*{PIF}
&3.7&72.3&2.3&83.9&3.9&83.1&1.3&82.9&2.7&80.8&1.3&93.0\\
~&~&
\textcolor{greenberry}{(\textbf{-6.6})} &\textcolor{red}{(+\textbf{1.4})}&
\textcolor{greenberry}{(\textbf{-4.5})} &\textcolor{red}{(+\textbf{2.1})}&
\textcolor{greenberry}{(\textbf{-2.3})} &\textcolor{red}{(+\textbf{1.8})}&
\textcolor{greenberry}{(\textbf{-6.8})} &\textcolor{red}{(+\textbf{2.7})}&
\textcolor{greenberry}{(\textbf{-5.0})} &\textcolor{red}{(+\textbf{1.9})}&
{(0)} &\textcolor{greenberry}{(-0.1)} \\ \hline

\multirow{5}*{LLaMA2-13B}&Symbol&6.2&73.9&1.9&84.9&5.6&79.5&3.5&83.9&3.8&80.5&2.8&92.5 \\ \cline{2-14}

~&\multirow{1}*{SCB}
&9.9&71.8&5.3&83.9&4.9&79.7&3.1&83.6&5.6&79.8&2.6&92.8\\

~&\multirow{1}*{RSCB}
&2.8&74.7&1.3&85.9&6.8&79.1&1.6&84.8&3.1&80.9&1.9&92.6 \\

~&\multirow{2}*{PIF}
&4.8&74.5&2.0&85.1&2.9&80.9&2.6&84.2&3.0&81.0&0.9&93.5\\
~&~&
\textcolor{greenberry}{(\textbf{-1.4})} &\textcolor{red}{(+\textbf{0.6})}&
\textcolor{red}{(+0.1)} &\textcolor{red}{(+0.2)}&
\textcolor{greenberry}{(\textbf{-2.7})} &\textcolor{red}{(+\textbf{1.4})}&
\textcolor{greenberry}{(\textbf{-0.9})} &\textcolor{red}{(+\textbf{0.3})}&
\textcolor{greenberry}{(\textbf{-0.8})} &\textcolor{red}{(+\textbf{0.5})}&
\textcolor{greenberry}{(\textbf{-1.9})} &\textcolor{red}{(+\textbf{1.0})} \\ 
\bottomrule

\end{tabular}
}
\end{center}
\caption{\label{table:mainresult}The metric $\mu_\text{bias}$ (A reduced value implies a diminished selection bias) and $\mu_\text{ppa}$ (An elevated value suggests an enhanced MSCB capability) of four methods. The implementation of PIF methodology has effectively enhanced the model's MCSB capability across virtually all datasets, consequently alleviating the selection bias.
}
\end{table*}
\setlength{\tabcolsep}{1.4pt}

\subsection{Point-wise Intelligent Feedback}
Human feedback allows LLMs to identify issues with accuracy, fairness, and bias~\citep{liu2023chain}.
Previous studies have explored how to incorporate human feedback with various stages during LLM's training process, such as pre-training \citep{korbak2023pretraining}, SFT \citep{yuan2023rrhf}, Reinforcement Learning from Human Feedback (RLHF)~\cite{stiennon2020learning,xue2023reinforcement}, and others.

In the context of MCQs, we possess knowledge of both the positive options' symbols and contents, as well as the negative options. We can also easily acquire negative symbol-content binding examples. We call this process \textit{Intelligent Feedback} without human preference annotations. Moreover, our feedback is intrinsically point-wise, i.e., with absolute scores, the reward score for positive samples should be $1$ and be $\lambda$ for negative samples. We refer to this approach as \textbf{P}oint-wise \textbf{I}ntelligent \textbf{F}eedback (PIF).
In this method, we aim to maximize the probability of positive examples, approaching 1, which can be optimized by cross-entropy loss, and minimize the likelihood of negative examples falling below $\lambda$.
Inspired by RRHF~\citep{yuan2023rrhf}, we optimize the object of negative samples as follows:

\begin{equation}
\begin{aligned}
\label{eq:neg}
\mathcal{L}_{\text{PIF}_\text{n}} & = \max(0, \\
                            &{\log \lambda}-\frac{\sum_t {\log} P_{\pi_\text{PIF}}(y^{\text{PIF}_\text{n}}_t|x,y^{\text{PIF}_\text{n}}_{<t})}{|y^{\text{PIF}_\text{n}}|}).
\end{aligned}
\end{equation}

Consequently, the overall optimization objective can be summarized as follows:
\begin{equation}
\begin{aligned}
\label{eq:PIF}
\mathcal{L}_{\text{PIF}} & = \max(0, -\gamma_k\cdot\\
                            &(\frac{{\sum_t {\log} P_{\pi_\text{PIF}}(y^{\text{PIF}}_t|x,y^{\text{PIF}}_{<t})}}{| y^{\text{PIF}} |}-\log \lambda_k)),
\end{aligned}
\end{equation}
where $k$ is utilized to differentiate between the positive and negative samples. In our experiments, 
$\gamma_\text{pos}=1$, 
$\gamma_\text{neg}=-1$, 
$\lambda_\text{pos}=1$,
$\lambda_\text{neg}$ is a hyper-parameter defined in Section~\ref{sec:implementation}.

We also find that the performance of $\pi_\text{PIF}$ is related to its initial parameters. We ultimately initialize $\pi_\text{PIF}$ using the parameters from $\pi_\text{RSCB}$. Appendix~\ref{sec:ablation study} demonstrates the ablation study.

\section{Experiment}
\subsection{Implementation Detail}
\label{sec:implementation}
\setlength{\tabcolsep}{4pt}
\begin{table*}

\begin{center}
\scalebox{0.86}{
\begin{tabular}{llcccccccccccc}
\toprule\noalign{\smallskip}
\multirow{2}*{\textbf{Model}}&\multirow{2}*{\textbf{Method}}&\multicolumn{2}{c}{\textbf{STEM}}&\multicolumn{2}{c}{\textbf{Social Science}}&\multicolumn{2}{c}{\textbf{Humanities}}&\multicolumn{2}{c}{\textbf{Others}}&\multicolumn{2}{c}{\textbf{MMLU}}&\multicolumn{2}{c}{\textbf{CSQA}} \\

\cmidrule(lr){3-4} \cmidrule(lr){5-6} \cmidrule(lr){7-8} \cmidrule(lr){9-10} \cmidrule(lr){11-12} \cmidrule(lr){13-14} 

~&~& Acc& $\text{Acc}_\text{min}$ & Acc&$\text{Acc}_\text{min}$& Acc&$\text{Acc}_\text{min}$& Acc&$\text{Acc}_\text{min}$& Acc&$\text{Acc}_\text{min}$& Acc&$\text{Acc}_\text{min}$\\ 
\noalign{\smallskip} \hline

\multirow{4}*{LLaMA2-7B}
 &Symbol   &
 43.8&33.5&
 62.5&54.5&
 52.2&43.3&
 60.6&51.6&
 54.6&45.6&
 \textbf{79.8}&
 77.2 \\
~&SCB  &
43.6&35.0&
62.1&55.9&
50.5&45.4&
60.4&54.9&
53.8&47.7&
78.4&76.2 \\
~&RSCB &
43.2&39.0&
62.3&56.8&
52.1&\textbf{45.5}&
60.5&56.8&
54.4&49.0&
79.7&74.9 \\
~&PIF  &
\textbf{44.0}&\textbf{39.5}&
\textbf{62.7}&\textbf{59.5}&
\textbf{53.1}&{42.4}&
\textbf{60.9}&\textbf{59.3}&
\textbf{55.0}&\textbf{49.5}&
79.4&\textbf{77.3} \\ \hline

\multirow{4}*{LLaMA2-13B}
 &Symbol   &
 46.9&42.3&
 68.1&64.0&
 55.0&47.9&
 65.5&59.9&
 59.2&54.8&
 81.5&76.2 \\ 
~&SCB  &
\textbf{48.7}&36.2&
68.5&62.0&
54.2&43.3&
64.8&61.8&
58.6&50.2&
81.0&79.0 \\
~&RSCB &
47.5&41.8&
68.0&65.2&
55.8&44.3&
64.6&61.8&
58.7&54.3&
80.9&77.9 \\
~&PIF  &
48.0&\textbf{42.4}&
\textbf{68.5}&\textbf{65.8}&
\textbf{58.4}&\textbf{52.2}&
\textbf{65.6}&\textbf{62.0}&
\textbf{60.0}&\textbf{56.8}&
\textbf{82.2}&\textbf{79.9} \\ 
\bottomrule

\end{tabular}
}
\end{center}
\caption{\label{table:acc}The Acc (accuracy of standard test dataset) and $\text{Acc}_\text{min}$ (Minimum accuracy after performing an answer-moving attack) of our four methods. 
By effectively constructing negative samples, the PIF method contributes to a notable increase in accuracy. Simultaneously, by alleviating the LLM's selection bias, it demonstrates a smaller decline in accuracy when faced with answer-moving attack, resulting in a higher $\text{Acc}_\text{min}$ compared to other methods.
}
\end{table*}
\setlength{\tabcolsep}{1.4pt}

\noindent\textbf{Samples For PIF}\quad
When employing PIF, the inclusion of negative samples in the optimization objective is essential. Given that the ultimate goal is to enhance the LLMs' MCSB performance, negative samples are constructed by randomly combining the incorrect option contents with all candidate option symbols. For instance, we can construct a negative sample "\textit{B: 33KV.}" for the example presented in Figure~\ref{fig:model backbone}. 
Due to the limited computational resources, we randomly select one negative sample for each instance.
What's more, in our experiment, the parameters of $\pi_\text{PIF}$ are warmed up from $\pi_\text{RSCB}$.

\noindent\textbf{Metrics}\quad
We finally employ four evaluation metrics. $\mu_\text{bias}$ is defined by Equation~\ref{eq:mubias} to measure the LLM's selection bias.
$\mu_\text{ppa}$ is defined by Equation~\ref{eq:mu_ppa} to represent the LLM's MCSB capability. 
"Acc" refers to the LLM's accuracy performance on the standard test set of the current benchmark.
$\text{Acc}_\text{min}$ is the minimum accuracy after performing the answer-moving attack, in which case LLMs with higher $\text{Acc}_\text{min}$ exhibit greater robustness.

Fine-tuning Hyper-parameters and prompts used in our experiment can be found in Appendix~\ref{sec:implementation details}.

\subsection{Main Results}
\begin{figure}[t]
\centering
\includegraphics[width=0.43\textwidth]{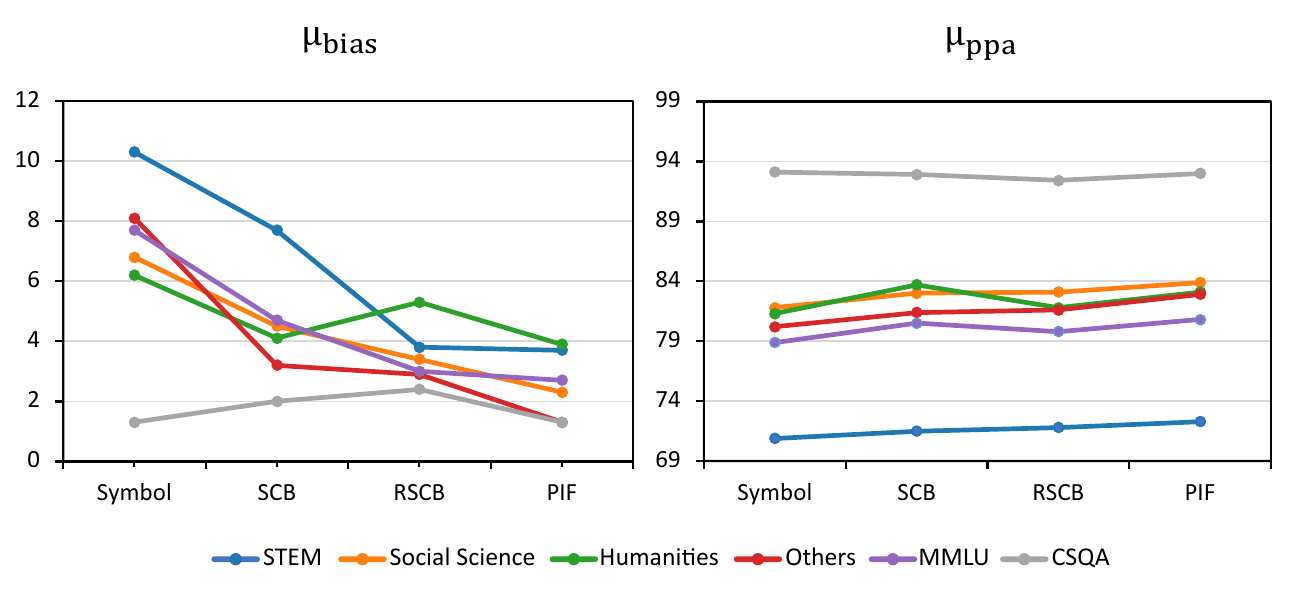}
\caption{
With the evolution of methods, $\mu_\text{bias}$ is gradually decreasing, $\mu_\text{ppa}$ is progressively increasing.}
\label{fig:zhexian}
\end{figure}

Table~\ref{table:mainresult} contains the main experimental results of comparison between four methods defined in Section~\ref{sec:method}, which are called "\textit{Symbol}", "\textit{Symbol-Content Binding (SCB)}", "\textit{Reweighting Symbol-Content Binding (RSCB)}" and "\textit{Point-wise Intelligent Feedback (PIF)}".
Figure~\ref{fig:zhexian} demonstrates a more apparent trend in the results' changes.
Appendix~\ref{sec:all results} also reports the accuracy after the answer-moving attack for each method on each benchmark.

The SCB method combines the option symbols and option contents as the prediction target tokens. However, this method can not reduce selection bias or enhance the LLM's MCSB capabilities across all datasets and LLMs. In fact, SCB can even exacerbate selection bias for LLaMA2-13B.

Conversely, RSCB balances the weight of option symbols and contents in the loss function based on SCB. This method enhances the abilities of MCSB on almost all datasets, which helps to reduce selection bias.
This finding implies that symbols play a more significant role in MCQs learning process.

Finally, we conduct experiments on our PIF method. This method further reduces selection bias for nearly all datasets. It is noteworthy that for LLaMA2-7B's performance on CSQA and LLaMA2-13B's performance on the Social Science benchmark, the inherent bias of the models is already relatively small, at $1.3$ and $1.9$, respectively. Under these circumstances, the bias of our PIF method is essentially on par with the bias of the \textit{Symbol} method. 
All these results validate the effectiveness of our constructed negative samples and the proposed point-wise objective function.

Moreover, based on the results presented in Table~\ref{table:mainresult}, it is evident that an enhancement in MCSB capability leads to a reduction in selection bias. This finding reaffirms our initial hypothesis:\textit{Strengthened Symbol Binding Makes Large Language Models Reliable Multiple-Choice Selectors}.

\subsection{Impact of Our Methods on Accuracy}
Essentially, while ensuring the stability of LLMs in MCQs, we also aspire to achieve higher accuracy.
In Table~\ref{table:acc}, we present the performance of four methods in terms of accuracy and also the minimum accuracy after the model has been subjected to an answer-moving attack. This aims to demonstrate whether the current method can guarantee accuracy when faced with adversarial attacks.

As we can see, the simple SCB method can not lead to an improvement in accuracy, and even causes a decrease in accuracy on LLaMA2-13B. 
On the other hand, by emphasizing the importance of symbols, RSCB achieves a similar level of accuracy as the method trained solely on symbols. Moreover, due to the significant reduction of selection bias in the RSCB method, it outperforms \textit{Symbol} in terms of the $\text{Acc}_\text{min}$ metric.

We are pleased to report that the PIF model outperforms the other three methods in terms of both Acc and $\text{Acc}_\text{min}$ metrics, yielding the most favorable results. This outstanding performance is attributed to the process of randomly selecting contents from incorrect options and combining them with all symbols while creating negative instances in the PIF model. 
This approach not only highlights incorrect binding relationships but also exposes the wrong option contents to the model, resulting in an enhancement of the model's accuracy.

\subsection{Discussion}
\noindent\textbf{What would happen if there were no bias in SFT training data?}
Although PIF has mitigated selection bias, there is a more aggressive debiasing approach. Considering that enhancing the model's MCSB capability can alleviate selection bias, similar to the calculation of $\mu_\text{ppa}$, we randomly combine symbols and contents to obtain all $K!$ possible arrangements as training data and train the model. We refer to this method as "\textit{Perm}".
Due to the limited computational resources, we randomly select $\frac{1}{K!}$ of the original training set for Perm.
The results are shown in Appendix~\ref{sec:permutation}. Perm indeed significantly alleviates selection bias, demonstrating that LLM's selection bias during SFT primarily stems from the training data utilized in the SFT. 
This is also consistent with the conclusion in Section~\ref{sec:selection bias}.

It is practically impossible to implement the \textit{Perm} approach as it requires using all possible random combinations, which is unrealistic due to the high computational resources and training time required. 
In contrast, PIF can effectively alleviate the selection bias by introducing just one negative example during the SFT process. This further confirms the effectiveness of our approach.

\noindent\textbf{PIF \textit{vs.} Data Argumentation.}
\textit{Perm} method can significantly reduce selection bias, but this approach becomes unfeasible due to its high computational resources. How about we only use a portion of the data in PIF for Data Augmentation? The resources consumed by this method are comparable to those of PIF. To verify the feasibility of this method, which is called \textit{Argum.}, we add an additional experiment based on LLaMA2-7B. Considering that PIF generates a random negative sample for each sample during training, we also augmented the original training data by randomly shuffling the combination of symbols and contents for each sample. The results are as Tabel~\ref{table:PIF vs argum}. As can be seen, under the same time consumption, simple data augmentation does not perform better than PIF. Since our PIF method is derived from RLHF, the rationale for its effectiveness is the same as that of RLHF, we introduce negative feedback information to the model, informing it that such negative examples cannot be generated, thereby improving the model's performance.

\noindent\textbf{Point-wise \textit{vs.} Pair-wise.}
We propose Point-wise Intelligent Feedback to resume the constructed negative samples. How about incorporating these feedback data using a Pair-wise method? 
We implement DPO~\citep{rafailov2023direct}, and the results in Appendix~\ref{sec:DPO} show that PIF outperforms both alleviating selection bias and achieving higher accuracy compared to DPO.
The results prove that it is rational to make the prediction scores of the negative examples approach a minimal value $\lambda$. However, when the samples can be optimized with absolute scores, the pair-wise method will lose the information of the gap between the pairs~\citep{cai2023ulma}.

\section{Related Work}

\textbf{Large Language Models}\quad In recent years, the scale of the model parameters has progressively increased with the rapid development of deep learning, from 1.5 billion in 2018 \citep{radford2018improving} to 540 billion in 2022 \citep{chowdhery2023palm}. A significant number of large language models have swiftly surfaced, especially after the emergence of chatGPT\citep{openai2022chatgpt}, such as LLaMA\citep{touvron2023llama}, Vicuna\citep{chiang2023vicuna}, and BLOOM\citep{workshop2022bloom}. These large language models have a vast amount of parameters. After being trained with massive data using generative methods, they possess the impressive ability of natural language understanding and can follow human instructions effectively \citep{ouyang2022training}.

\noindent \textbf{Selection Bias in Multiple-Choice Questions}\quad The robustness and vulnerabilities of large language models have always been an important research realm. Many researchers have explored how large language models are influenced by modifications or adversarial attacks that impact individual instances in few-shot learning. For example, \citet{zhao2021calibrate} found that large language models are easily affected by changes in task instructions and context when performing tasks. \citet{Wang2023LargeLM} and \cite{zheng2023judging} reveal that GPT-4 tends to choose the option in the first position.

MCQs are widely used to assess the capabilities of large language models. Various MCQs datasets are introduced as standard language model benchmarks, such as MMLU\citep{hendryckstest2021}, C-Eval\citep{huang2023c}, CSQA\citep{talmor-etal-2019-commonsenseqa}. LLMs have achieved human-like performances on various MCQ benchmarks. However, many researchers have noticed that LLMs suffer from selection bias in MCQS.
\citet{pezeshkpour2023large} shows that large language models are sensitive to the order of choices. \citet{zheng2023large} found that LLMs are vulnerable to option symbol changes in MCQs due to their inherent "token bias". \citet{robinson2023iclr} shows that a model with high multiple choice symbol binding(MCSB) ability performs better in MCQs.
However, previous studies have only explored the selection bias of LLMs in few-shot scenarios. 
Our work demonstrates that the selection bias still exists in SFT and mainly stems from the LLM’s inadequate multiple choice symbol binding capability.


\noindent \textbf{Human Feedback in LLMs}\quad \citet{openai2022chatgpt} and  \citet{schulman2017proximal} have demonstrated the potential of applying reinforcement learning in LLMs and its impressive performance. One part of the main techniques is learning a reward function from human feedback for reinforcement learning, which is often dubbed as RLHF\citep{christiano2017deep} \citep{macglashan2017interactive} \citep{lee2021pebble}. However, RLHF is a complex, expensive, and often unstable procedure. \citet{lee2023rlaif} offers a promising alternative that uses a powerful LLM to generate preference instead of human annotators. To mitigate the problem of PPO's sizeable computational cost, \citet{rafailov2023direct} introduces a stable, computationally lightweight method to solve the standard RLHF problem. \citet{liu2023chain} convert all types of feedback into sequences of sentences, which are then used to fine-tune the model to learn human preference. The methods above mainly focus on pair-wise preference data. To use point-wise preference data, \citet{cai2023ulma} develop a point-wise preference learning method. In this paper, we propose PIF which designs a point-wise loss to incorporate negative samples into SFT.








\section{Conclusion}
This paper highlights that selection bias persists in the SFT of LLMs in MCQs. To explore why LLMs suffer from selection bias, we suppose that "Symbol Binding Makes Large Language Models Reliable Multiple-Choice Selectors". 
We utilize MCSB capability to represent the LLMs' ability to bind option symbols and the corresponding content and design experiments to establish the relationship between MCSB capability and the selection bias.
Finally, we eliminate selection bias by enhancing the model's MCSB capability. 
We propose PIF, constructing negative samples by randomly combining the content of incorrect contents with all candidate symbols and designing a point-wise loss to resume these negative samples.
Comprehensive experimental results demonstrate that PIF significantly reduces selection bias and substantially improves the accuracy of LLMs for MCQs.

\section{Limitations}
Due to limited computational resources, there are some deficiencies in our work and we list here for future reference. 
Firstly, we conduct experiments on LLMs with 7B and 13B parameter sizes. In the future, we are eager to understand the patterns of selection bias during the SFT phase in LLMs with even larger parameter sizes such as 70B.
Secondly, during our experiments, we select one negative sample randomly for each instance. 
We would like to investigate the correlation between the severity of selection bias and the percentage of negative samples introduced. 
Finally, our method has a broader range of applications.  For instance,  when using LLMs as human preference annotators, they prefer the responses in a specific position. In this case, we can assign some symbols to all responses and then use PIF method to eliminate the bias.
We leave these limitations for future work.

\bibliography{custom}
\clearpage

\appendix

\section{Statistics of Benchmarks}

\label{sec:benchmark statistics}
Statistics of all benchmarks used in our paper is illustrated in Table~\ref{table:data_statics}.

\section{LLMs Project}
\label{sec:llms project}

\textbf{LLaMA2-7B}\quad https://huggingface.co/meta-llama/Llama-2-7b-hf

\noindent\textbf{LLaMA2-13B}\quad https://huggingface.co/meta-llama/Llama-2-13b-hf

\noindent\textbf{Implementation Framework}\quad 

\noindent https://github.com/hiyouga/LLaMA-Factory

\section{Implementation Details}
\label{sec:implementation details}
\noindent\textbf{Fine-tuning Hyper-parameters}\quad
We assign $\alpha$ and $\beta$ of $\mathcal{L}_{\text{RSCB}}$ to $2$ and $0.1$, respectively.
For $\lambda$ of $\mathcal{L}_{\text{PIF}}$, we conduct experiments with a set of values $\{0.0001, 0.001, 0.01, 0.1\}$, and select the one that achieves the best performance on the validation set.
We configure the sequence length, epoch, the maximum number of new tokens generated as 1024, 3, 4, respectively. 
For learning rate, we experiment with the value set $\{1e-5, 5e-5, 1e-4, 2e-4\}$, and select the one that yielded the best performance on validation. In most cases, it is set to $1e-4$.
We conduct experiments on 8 GPUs. 
For the MMLU benchmark, for LLaMA2-7B and LLaMA2-13B models, the batch size per device is set to 4 and 2, respectively, while the gradient accumulation step is set to 2 and 4, ensuring a final total batch size of 64. For CSQA, we adjust the gradient accumulation step to achieve a final total batch size of 32.
Additional details can be found in our code.

\noindent\textbf{Prompts}\quad
We don't use specific prompt engineering in our experiments. Instead, all the experiments are conducted using a simple prompt:

\noindent"\textit{The following are multiple choice questions. You should directly answer the question by choosing the correct option.}

\noindent\textit{Question: \{\{text\}\}}

\noindent\textit{Options: \{\{text\}\}}

\noindent\textit{Answer: \{\{text\}\}}"

\section{Ablation Study on PIF Parameter Initialization}
\label{sec:ablation study}
In this part, we investigate the effectiveness of the parameter initialization for $\pi_\text{PIF}$, the results are contained in Table~\ref{table:ablationstudy}. 
Compared to $\text{PIF}_{\text{raw}}$, our proposed $\text{PIF}_{\text{RSCB}}$ performs better in terms of selection bias, MSCB capability and accuracy.

\section{Results of Permutation}

\begin{figure*}[t]
\centering
\includegraphics[width=0.9\textwidth]{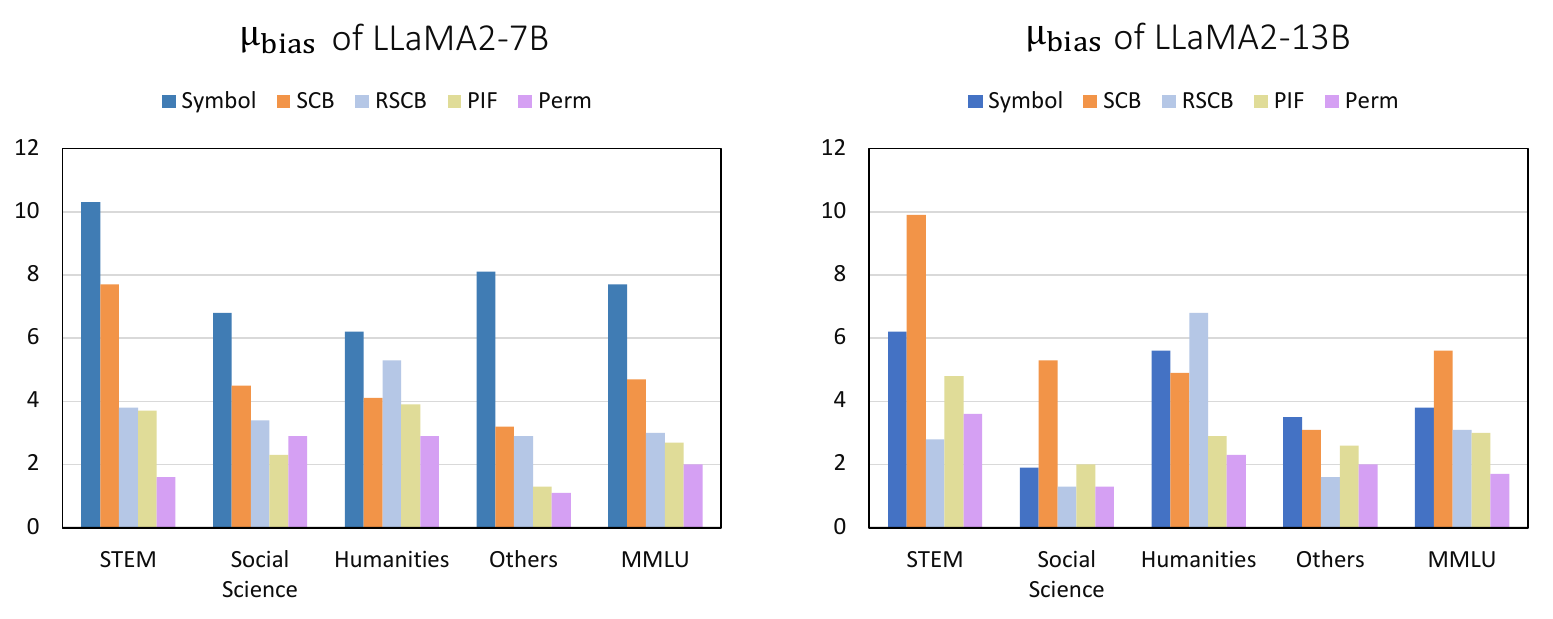}
\caption{Selection bias of Perm.}
\label{fig:permutation}
\end{figure*}

Figure~\ref{fig:permutation} illustrates the selection bias of Perm.
Perm indeed significantly alleviates selection bias, which demonstrates that LLM's selection bias during SFT primarily stems from the training data for SFT.

\section{Results of The Point-wise and Pair-wise Method}
\label{sec:DPO}
Table~\ref{table:pairwise vs pointwise} demonstrates the different results between point-wise method PIF and pair-wise method DPO. PIF outperforms both alleviating selection bias and achieving higher accuracy compared to DPO.

\setlength{\tabcolsep}{4pt}
\begin{table*}[t]
\begin{center}
\scalebox{0.88}{
\begin{tabular}{ccccccc}
\toprule
\multicolumn{3}{l}{\textbf{Benchmarks}} & \textbf{\#Samples}  & \textbf{\#Options} & \textbf{Golden Answer Distribution} \\ 
\noalign{\smallskip}\hline
 \multirow{6}*{MMLU} 
    & train & -  & 99842   & 4 & 22.2\%/25.8\%/26.9\%/25.1\% \\ 
    \cline{2-7}
  ~ & \multirow{5}*{test} & STEM & 3018 & \multirow{5}*{4} & 21.4\%/23.8\%/25.9\%/28.9\% \\
  ~ & ~ & Social Science & 3077 & ~ & 21.7\%/23.4\%/23.8\%/31.1\% \\
  ~ & ~ & Humanities & 4705 & ~ & 24.2\%/24.5\%/27.1\%/24.2\% \\
  ~ & ~ & Others & 3242 & ~ & 23.8\%/26.8\%/24.4\%/25.1\% \\ 
  ~ & ~ & Overall & 14042 & ~ &  22.3\%/24.7\%/25.5\%/26.9\% \\ \hline
\multirow{2}*{CSQA}
   & \multicolumn{2}{c}{train} & 8971 & \multirow{2}*{5}  & 19.4\%/20.3\%/20.1\%/20.4\%/19.9\% &  \\ 
 ~ & \multicolumn{2}{c}{test} & 1221 & ~ & 19.6\%/20.9\%/19.7\%/20.6\%19.3\% &  \\ 
\bottomrule
\end{tabular}
}
\end{center}
\caption{\label{table:data_statics}Statistics of all benchmarks.}
\end{table*}
\setlength{\tabcolsep}{1.4pt}

\setlength{\tabcolsep}{4pt}
\begin{table*}
\begin{center}
\scalebox{0.74}{
\begin{tabular}{ll|ccc|ccc|ccc|ccc|ccc}
\hline
\multicolumn{1}{c}{\multirow{2}*{\textbf{Model}}}
&\multicolumn{1}{c|}{\multirow{2}*{\textbf{Method}}}
&\multicolumn{3}{c|}{\textbf{STEM}}
&\multicolumn{3}{c|}{\textbf{Social Science}}
&\multicolumn{3}{c|}{\textbf{Humanities}}
&\multicolumn{3}{c|}{\textbf{Others}}
&\multicolumn{3}{c}{\textbf{MMLU}} 
\\ \cline{3-17}
~&~&$\mu_\text{bias}$$\downarrow$&$\mu_\text{ppa}$$\uparrow$&Acc$\uparrow$
&$\mu_\text{bias}$$\downarrow$&$\mu_\text{ppa}$$\uparrow$&Acc$\uparrow$
&$\mu_\text{bias}$$\downarrow$&$\mu_\text{ppa}$$\uparrow$&Acc$\uparrow$
&$\mu_\text{bias}$$\downarrow$&$\mu_\text{ppa}$$\uparrow$&Acc$\uparrow$
&$\mu_\text{bias}$$\downarrow$&$\mu_\text{ppa}$$\uparrow$&Acc$\uparrow$ \\ \hline
\multirow{4}*{LLaMA2-7B}
&$\text{PIF}_{\text{raw}}$&9.6&67.4&43.5&6.2&78.8&61.4&10.1&70.1&48.7&4.9&79.0&59.2&7.9&73.5&52.8 \\
&$\text{PIF}_{\text{Symbol}}$&8.3&70.6&44.1&2.6&83.6&\textbf{63.4}&4.7&82.9&51.8&4.5&81.6&\textbf{61.1}&3.7&79.6&54.8 \\
&$\text{PIF}_{\text{SCB}}$&6.9&68.4&41.5&3.6&81.0&60.4&\textbf{2.6}&82.4&49.0&6.1&78.8&59.0&4.4&78.3&52.2 \\
~&$\text{PIF}_{\text{RSCB}}$&\textbf{3.7}&\textbf{72.3}&\textbf{44.0}&\textbf{2.3}&\textbf{83.9}&62.7&3.9&\textbf{83.1}&\textbf{53.1}&\textbf{1.3}&\textbf{82.9}&60.9&\textbf{2.7}&\textbf{80.8}&\textbf{55.0} \\
\hline

\multirow{4}*{LLaMA2-13B}
&$\text{PIF}_{\text{raw}}$&5.1&73.4&47.9&2.3&84.7&69.4&7.2&78.1&54.2&2.2&84.5&65.2&3.6&80.1&58.7 \\
&$\text{PIF}_{\text{Symbol}}$&\textbf{3.2}&\textbf{75.1}&46.2&4.0&83.6&66.6&4.9&78.6&56.2&3.6&83.9&62.7&6.5&80.4&57.8 \\
&$\text{PIF}_{\text{SCB}}$&4.5&73.6&47.5&2.2&84.2&68.0&4.7&80.3&58.0&\textbf{2.1}&\textbf{84.9}&64.7&3.2&81.0&59.5 \\
~&$\text{PIF}_{\text{RSCB}}$&4.8&74.5&\textbf{48.0}&\textbf{2.0}&\textbf{85.1}&\textbf{68.5}&\textbf{2.9}&\textbf{80.9}&\textbf{58.4}&2.6&84.2&\textbf{65.5}&\textbf{3.0}&\textbf{81.0}&\textbf{60.0} \\
\hline
\end{tabular}
}
\end{center}
\caption{\label{table:ablationstudy} Ablation study on $\pi_\text{PIF}$ parameter initialization. $\text{PIF}_{\text{raw}}$ refers to the model $\pi_\text{PIF}$ which is initialized from the models listed in Appendix~\ref{sec:llms project}. $\text{PIF}_{\text{Symbol}}$, $\text{PIF}_{\text{SCB}}$, $\text{PIF}_{\text{RSCB}}$ refers to the model $\pi_\text{PIF}$ which is initialized from $\pi_\text{Symbol}$, $\pi_\text{SCB}$, $\pi_\text{RSCB}$ respectively. }
\end{table*}
\setlength{\tabcolsep}{1.4pt}

\setlength{\tabcolsep}{4pt}
\begin{table*}
\begin{center}
\scalebox{0.85}{
\begin{tabular}{ll|cc|cc|cc|cc|cc}
\hline
\multicolumn{1}{c}{\multirow{2}*{\textbf{Model}}}
&\multicolumn{1}{c|}{\multirow{2}*{\textbf{Method}}}
&\multicolumn{2}{c|}{\textbf{STEM}}
&\multicolumn{2}{c|}{\textbf{Social Science}}
&\multicolumn{2}{c|}{\textbf{Humanities}}
&\multicolumn{2}{c|}{\textbf{Others}}
&\multicolumn{2}{c}{\textbf{MMLU}} 
\\ \cline{3-12}
~&~&$\mu_\text{bias}$$\downarrow$&Acc$\uparrow$
&$\mu_\text{bias}$$\downarrow$&Acc$\uparrow$
&$\mu_\text{bias}$$\downarrow$&Acc$\uparrow$
&$\mu_\text{bias}$$\downarrow$&Acc$\uparrow$
&$\mu_\text{bias}$$\downarrow$&Acc$\uparrow$ \\ \hline
\multirow{2}*{LLaMA2-7B}
&Argum.&\textbf{2.0}&\textbf{44.1}&2.8&62.7&5.9&51.1&1.5&60.4&2.7&54.3 \\
~&PIF&3.7&44.0&\textbf{2.3}&\textbf{62.7}&\textbf{3.9}&\textbf{53.1}&\textbf{1.3}&\textbf{60.9}&\textbf{2.7}&\textbf{55.0} \\
\hline
\end{tabular}
}
\end{center}
\caption{\label{table:PIF vs argum} PIF \textit{vs.} Data Argumentation.}
\end{table*}
\setlength{\tabcolsep}{1.4pt}

\setlength{\tabcolsep}{4pt}
\begin{table*}
\begin{center}
\scalebox{0.73}{
\begin{tabular}{ll|ccc|ccc|ccc|ccc|ccc}
\hline
\multicolumn{1}{c}{\multirow{2}*{\textbf{Model}}}
&\multicolumn{1}{c|}{\multirow{2}*{\textbf{Method}}}
&\multicolumn{3}{c|}{\textbf{STEM}}
&\multicolumn{3}{c|}{\textbf{Social Science}}
&\multicolumn{3}{c|}{\textbf{Humanities}}
&\multicolumn{3}{c|}{\textbf{Others}}
&\multicolumn{3}{c}{\textbf{MMLU}} 
\\ \cline{3-17}
~&~&$\mu_\text{bias}$$\downarrow$&$\mu_\text{ppa}$$\uparrow$&Acc$\uparrow$
&$\mu_\text{bias}$$\downarrow$&$\mu_\text{ppa}$$\uparrow$&Acc$\uparrow$
&$\mu_\text{bias}$$\downarrow$&$\mu_\text{ppa}$$\uparrow$&Acc$\uparrow$
&$\mu_\text{bias}$$\downarrow$&$\mu_\text{ppa}$$\uparrow$&Acc$\uparrow$
&$\mu_\text{bias}$$\downarrow$&$\mu_\text{ppa}$$\uparrow$&Acc$\uparrow$ \\ \hline
\multirow{2}*{LLaMA2-7B}
&DPO&8.9&70.6&43.1&3.0&80.4&62.6&3.9&77.2&52.8&5.1&81.6&59.9&5.0&77.5&54.5 \\
~&PIF&\textbf{3.7}&\textbf{72.3}&\textbf{44.0}&\textbf{2.3}&\textbf{83.9}&\textbf{62.7}&\textbf{3.9}&\textbf{83.1}&\textbf{53.1}&\textbf{1.3}&\textbf{82.9}&\textbf{60.9}&\textbf{2.7}&\textbf{80.8}&\textbf{55.0} \\
\hline

\multirow{2}*{LLaMA2-13B}
&DPO&7.6&69.6&47.4&3.8&83.1&67.0&7.5&78.3&58.4&5.5&81.6&63.9&4.9&78.2&58.4 \\
~&PIF&\textbf{4.8}&\textbf{74.5}&\textbf{48.0}&\textbf{2.0}&\textbf{85.1}&\textbf{68.5}&\textbf{2.9}&\textbf{80.9}&\textbf{58.4}&\textbf{2.6}&\textbf{84.2}&\textbf{65.5}&\textbf{3.0}&\textbf{81.0}&\textbf{60.0} \\
\hline
\end{tabular}
}
\end{center}
\caption{\label{table:pairwise vs pointwise} Point-wise \textit{vs.} Pair-wise.}
\end{table*}
\setlength{\tabcolsep}{1.4pt}

\section{Evaluation Results of Selection Bias}
\label{sec:all results}
All accuracy results after answer-moving attack can be found in Table~\ref{table:STEM results},~\ref{table:Others results},~\ref{table:Social results},~\ref{table:Humanities results},~\ref{table:MMLU results},~\ref{table:CSQA results}.

\setlength{\tabcolsep}{4pt}
\begin{table}[t]
\begin{center}
\scalebox{0.7}{
\begin{tabular}{lcccccc}
\toprule
\multicolumn{2}{c}{\textbf{Move Golden to}}  & \textbf{orig} & \textbf{A} & \textbf{B} & \textbf{C} & \textbf{D}\\ 
\noalign{\smallskip} \hline

 \multirow{8}*{LLaMA2-7B} 
   & \multirow{2}*{Symbol} & \multirow{2}*{43.8} & 61.6 & 33.9 & 47.2 & 33.5  \\ 
   ~ & ~ & ~ & \textcolor{red}{(+17.8)} & \textcolor{greenberry}{(-9.9)} & \textcolor{red}{(+3.4)} & \textcolor{greenberry}{(-10.3)}  \\
   & \multirow{2}*{SCB} & \multirow{2}*{43.6} & 35.0 & 35.4 & 52.4 & 48.7  \\ 
   ~ & ~ & ~ & \textcolor{greenberry}{(-8.6)} & \textcolor{greenberry}{(-8.2)} & \textcolor{red}{(+8.8)} & \textcolor{red}{(+5.1)}  \\ 
   & \multirow{2}*{RSCB} & \multirow{2}*{43.2} & 47.6 & 40.7 & 39.0 & 47.4  \\ 
   ~ & ~ & ~ & \textcolor{red}{(+4.4)} & \textcolor{greenberry}{(-2.5)} & \textcolor{greenberry}{(-4.2)} & \textcolor{red}{(+4.2)}  \\  
   & \multirow{2}*{PIF} & \multirow{2}*{44.0} & 39.5 & 45.8 & 40.6 & 49.2  \\ 
   ~ & ~ & ~ & \textcolor{greenberry}{(-4.5)} & \textcolor{red}{(+1.8)} & \textcolor{greenberry}{(-3.4)} & \textcolor{red}{(+5.2)}  \\  \hline
 \multirow{8}*{LLaMA2-13B} 
   & \multirow{2}*{Symbol} & \multirow{2}*{46.9} & 43.4 & 57.4 & 42.3 & 53.4  \\ 
   ~ & ~ & ~ & \textcolor{greenberry}{(-3.5)} & \textcolor{red}{(+10.5)} & \textcolor{greenberry}{(-4.6)} & \textcolor{red}{(+6.5)}  \\
   & \multirow{2}*{SCB} & \multirow{2}*{48.7} & 36.2 & 41.9 & 45.2 & 65.7  \\ 
   ~ & ~ & ~ & \textcolor{greenberry}{(-12.5)} & \textcolor{greenberry}{(-6.8)} & \textcolor{greenberry}{(-3.5)} & \textcolor{red}{(+17.0)}  \\  
   & \multirow{2}*{RSCB} & \multirow{2}*{47.5} & 47.3 & 51.2 & 49.2 & 41.8  \\ 
   ~ & ~ & ~ & \textcolor{greenberry}{(-0.2)} & \textcolor{red}{(+3.7)} & \textcolor{red}{(+1.7)} & \textcolor{greenberry}{(-5.7)}  \\  
  & \multirow{2}*{PIF} & \multirow{2}*{48.0} & 47.5 & 45.7 & 58.8 & 42.4  \\ 
   ~ & ~ & ~ & \textcolor{greenberry}{(-0.5)} & \textcolor{greenberry}{(-2.3)} & \textcolor{red}{(+10.8)} & \textcolor{greenberry}{(-5.6)}  \\  
\bottomrule
\end{tabular}
}
\end{center}
\caption{\label{table:STEM results}The accuracy results after answer-moving attack on the LLMs with STEM benchmarks.}
\end{table}
\setlength{\tabcolsep}{1.4pt}

\setlength{\tabcolsep}{4pt}
\begin{table}[t]
\begin{center}
\scalebox{0.7}{
\begin{tabular}{lcccccc}
\toprule
\multicolumn{2}{c}{\textbf{Move Golden to}}  & \textbf{orig} & \textbf{A} & \textbf{B} & \textbf{C} & \textbf{D}\\ 
\noalign{\smallskip} \hline

 \multirow{8}*{LLaMA2-7B} 
   & \multirow{2}*{Symbol} & \multirow{2}*{60.6} & 73.8 & 51.6 & 64.2 & 53.9  \\ 
   ~ & ~ & ~ & \textcolor{red}{(+13.2)} & \textcolor{greenberry}{(-9.0)} & \textcolor{red}{(+3.6)} & \textcolor{greenberry}{(-6.7)}  \\
   & \multirow{2}*{SCB} & \multirow{2}*{60.4} & 58.5 & 54.9 & 63.6 & 62.7  \\ 
   ~ & ~ & ~ & \textcolor{greenberry}{(-1.9)} & \textcolor{greenberry}{(-5.5)} & \textcolor{red}{(+3.2)} & \textcolor{red}{(+2.3)}  \\ 
   & \multirow{2}*{RSCB} & \multirow{2}*{60.5} & 62.1 & 58.9 & 56.8 & 65.0  \\ 
   ~ & ~ & ~ & \textcolor{red}{(+1.6)} & \textcolor{greenberry}{(-1.6)} & \textcolor{greenberry}{(-3.7)} & \textcolor{red}{(+4.5)}  \\  
   & \multirow{2}*{PIF} & \multirow{2}*{60.9} & 59.3 & 60.7 & 59.3 & 62.9  \\ 
   ~ & ~ & ~ & \textcolor{greenberry}{(-1.6)} & \textcolor{greenberry}{(-0.2)} & \textcolor{greenberry}{(-1.6)} & \textcolor{red}{(+2.0)}  \\  \hline
 \multirow{8}*{LLaMA2-13B} 
   & \multirow{2}*{Symbol} & \multirow{2}*{65.5} & 64.3 & 70.5 & 59.9 & 63.4  \\ 
   ~ & ~ & ~ & \textcolor{greenberry}{(-1.2)} & \textcolor{red}{(+5.0)} & \textcolor{greenberry}{(-5.6)} & \textcolor{greenberry}{(-2.1)}  \\
   & \multirow{2}*{SCB} & \multirow{2}*{64.8} & 62.1 & 63.8 & 61.8 & 70.5  \\ 
   ~ & ~ & ~ & \textcolor{greenberry}{(-2.7)} & \textcolor{greenberry}{(-1.0)} & \textcolor{greenberry}{(-3.0)} & \textcolor{red}{(+5.7)}  \\  
   & \multirow{2}*{RSCB} & \multirow{2}*{64.6} & 65.0 & 66.3 & 66.0 & 61.8  \\ 
   ~ & ~ & ~ & \textcolor{red}{(+0.4)} & \textcolor{red}{(+1.7)} & \textcolor{red}{(+1.4)} & \textcolor{greenberry}{(-2.8)}  \\  
  & \multirow{2}*{PIF} & \multirow{2}*{65.6} & 64.7 & 63.8 & 70.1 & 62.0  \\ 
   ~ & ~ & ~ & \textcolor{greenberry}{(-0.9)} & \textcolor{greenberry}{(-1.8)} & \textcolor{red}{(+4.5)} & \textcolor{greenberry}{(-3.6)}  \\  
\bottomrule
\end{tabular}
}
\end{center}
\caption{\label{table:Others results}The accuracy results after answer-moving attack on the LLMs with Others benchmarks.}
\end{table}
\setlength{\tabcolsep}{1.4pt}

\setlength{\tabcolsep}{4pt}
\begin{table}[t]
\begin{center}
\scalebox{0.65}{
\begin{tabular}{lcccccc}
\toprule
\multicolumn{2}{c}{\textbf{Move Golden to}}  & \textbf{orig} & \textbf{A} & \textbf{B} & \textbf{C} & \textbf{D}\\ 
\noalign{\smallskip} \hline

 \multirow{8}*{LLaMA2-7B} 
   & \multirow{2}*{Symbol} & \multirow{2}*{62.5} & 73.0 & 54.5 & 64.6 & 55.9  \\ 
   ~ & ~ & ~ & \textcolor{red}{(+10.5)} & \textcolor{greenberry}{(-8.0)} & \textcolor{red}{(+2.1)} & \textcolor{greenberry}{(-6.6)}  \\
   & \multirow{2}*{SCB} & \multirow{2}*{62.1} & 56.8 & 55.9 & 65.2 & 65.4  \\ 
   ~ & ~ & ~ & \textcolor{greenberry}{(-5.3)} & \textcolor{greenberry}{(-6.2)} & \textcolor{red}{(+3.1)} & \textcolor{red}{(+3.3)}  \\ 
   & \multirow{2}*{RSCB} & \multirow{2}*{62.3} & 61.2 & 59.4 & 56.8 & 66.6  \\ 
   ~ & ~ & ~ & \textcolor{greenberry}{(-1.1)} & \textcolor{greenberry}{(-2.9)} & \textcolor{greenberry}{(-5.5)} & \textcolor{red}{(+4.3)}  \\  
   & \multirow{2}*{PIF} & \multirow{2}*{62.7} & 59.5 & 60.5 & 61.1 & 65.2  \\ 
   ~ & ~ & ~ & \textcolor{greenberry}{(-3.2)} & \textcolor{greenberry}{(-2.2)} & \textcolor{greenberry}{(-1.6)} & \textcolor{red}{(+2.5)}  \\  \hline
 \multirow{8}*{LLaMA2-13B} 
   & \multirow{2}*{Symbol} & \multirow{2}*{68.1} & 66.7 & 70.0 & 64.0 & 67.8  \\ 
   ~ & ~ & ~ & \textcolor{greenberry}{(-1.4)} & \textcolor{red}{(+1.9)} & \textcolor{greenberry}{(-4.1)} & \textcolor{greenberry}{(-0.3)}  \\
   & \multirow{2}*{SCB} & \multirow{2}*{68.5} & 62.0 & 64.7 & 63.6 & 74.7  \\ 
   ~ & ~ & ~ & \textcolor{greenberry}{(-6.5)} & \textcolor{greenberry}{(-3.8)} & \textcolor{greenberry}{(-4.9)} & \textcolor{red}{(+6.2)}  \\  
   & \multirow{2}*{RSCB} & \multirow{2}*{68.0} & 65.2 & 67.8 & 69.9 & 67.8  \\ 
   ~ & ~ & ~ & \textcolor{greenberry}{(-2.8)} & \textcolor{greenberry}{(-0.2)} & \textcolor{red}{(+1.9)} & \textcolor{greenberry}{(-0.2)}  \\  
  & \multirow{2}*{PIF} & \multirow{2}*{68.5} & 67.3 & 65.8 & 70.0 & 65.8  \\ 
   ~ & ~ & ~ & \textcolor{greenberry}{(-1.2)} & \textcolor{greenberry}{(-2.7)} & \textcolor{red}{(+1.5)} & \textcolor{greenberry}{(-2.7)}  \\  
\bottomrule
\end{tabular}
}
\end{center}
\caption{\label{table:Social results}The accuracy results after answer-moving attack on the LLMs with Social science benchmarks.}
\end{table}
\setlength{\tabcolsep}{1.4pt}

\setlength{\tabcolsep}{4pt}
\begin{table}[t]
\begin{center}
\scalebox{0.7}{
\begin{tabular}{lcccccc}
\toprule
\multicolumn{2}{c}{\textbf{Move Golden to}}  & \textbf{orig} & \textbf{A} & \textbf{B} & \textbf{C} & \textbf{D}\\ 
\noalign{\smallskip} \hline

 \multirow{8}*{LLaMA2-7B} 
   & \multirow{2}*{Symbol} & \multirow{2}*{52.2} & 58.0 & 43.3 & 57.4 & 47.7  \\ 
   ~ & ~ & ~ & \textcolor{red}{(+5.8)} & \textcolor{greenberry}{(-8.9)} & \textcolor{red}{(+5.2)} & \textcolor{greenberry}{(-4.5)}  \\
   & \multirow{2}*{SCB} & \multirow{2}*{50.5} & 45.9 & 45.4 & 54.0 & 53.7  \\ 
   ~ & ~ & ~ & \textcolor{greenberry}{(-4.6)} & \textcolor{greenberry}{(-5.1)} & \textcolor{red}{(+3.5)} & \textcolor{red}{(+3.2)}  \\ 
   & \multirow{2}*{RSCB} & \multirow{2}*{52.1} & 46.1 & 54.0 & 45.5 & 58.4  \\ 
   ~ & ~ & ~ & \textcolor{greenberry}{(-6.0)} & \textcolor{red}{(+1.9)} & \textcolor{greenberry}{(-6.6)} & \textcolor{red}{(+6.3)}  \\  
   & \multirow{2}*{PIF} & \multirow{2}*{53.1} & 42.4 & 53.0 & 51.5 & 56.7  \\ 
   ~ & ~ & ~ & \textcolor{greenberry}{(-10.7)} & \textcolor{greenberry}{(-0.1)} & \textcolor{greenberry}{(-1.6)} & \textcolor{red}{(+3.6)}  \\  \hline
 \multirow{8}*{LLaMA2-13B} 
   & \multirow{2}*{Symbol} & \multirow{2}*{55.0} & 47.9 & 61.6 & 57.4 & 61.2  \\ 
   ~ & ~ & ~ & \textcolor{greenberry}{(-7.1)} & \textcolor{red}{(+6.6)} & \textcolor{red}{(+2.4)} & \textcolor{red}{(+6.2)}  \\
   & \multirow{2}*{SCB} & \multirow{2}*{54.2} & 43.3 & 55.8 & 55.9 & 65.5  \\ 
   ~ & ~ & ~ & \textcolor{greenberry}{(-10.9)} & \textcolor{red}{(+1.6)} & \textcolor{red}{(+1.7)} & \textcolor{red}{(+11.3)}  \\  
   & \multirow{2}*{RSCB} & \multirow{2}*{55.8} & 44.3 & 61.1 & 64.3 & 57.9  \\ 
   ~ & ~ & ~ & \textcolor{greenberry}{(-11.5)} & \textcolor{red}{(+5.3)} & \textcolor{red}{(+8.5)} & \textcolor{red}{(+2.1)}  \\  
  & \multirow{2}*{PIF} & \multirow{2}*{58.4} & 52.2 & 58.9 & 61.6 & 56.6  \\ 
   ~ & ~ & ~ & \textcolor{greenberry}{(-6.2)} & \textcolor{red}{(+0.5)} & \textcolor{red}{(+3.2)} & \textcolor{greenberry}{(-1.8)}  \\  
\bottomrule
\end{tabular}
}
\end{center}
\caption{\label{table:Humanities results}The accuracy results after answer-moving attack on the LLMs with Humanities benchmarks.}
\end{table}
\setlength{\tabcolsep}{1.4pt}

\setlength{\tabcolsep}{4pt}
\begin{table}[t]
\begin{center}
\scalebox{0.7}{
\begin{tabular}{lcccccc}
\toprule
\multicolumn{2}{c}{\textbf{Move Golden to}}  & \textbf{orig} & \textbf{A} & \textbf{B} & \textbf{C} & \textbf{D}\\ 
\noalign{\smallskip} \hline

 \multirow{8}*{LLaMA2-7B} 
   & \multirow{2}*{Symbol} & \multirow{2}*{54.6} & 65.7 & 45.6 & 58.4 & 47.8  \\ 
   ~ & ~ & ~ & \textcolor{red}{(+11.1)} & \textcolor{greenberry}{(-9.0)} & \textcolor{red}{(+3.8)} & \textcolor{greenberry}{(-6.8)}  \\
   & \multirow{2}*{SCB} & \multirow{2}*{53.8} & 48.9 & 47.7 & 58.3 & 57.2  \\ 
   ~ & ~ & ~ & \textcolor{greenberry}{(-4.9)} & \textcolor{greenberry}{(-6.1)} & \textcolor{red}{(+4.5)} & \textcolor{red}{(+3.4)}  \\ 
   & \multirow{2}*{RSCB} & \multirow{2}*{54.4} & 53.5 & 53.5 & 49.0 & 59.3  \\ 
   ~ & ~ & ~ & \textcolor{greenberry}{(-0.9)} & \textcolor{greenberry}{(-0.9)} & \textcolor{greenberry}{(-5.4)} & \textcolor{red}{(+4.9)}  \\  
   & \multirow{2}*{PIF} & \multirow{2}*{55.0} & 49.5 & 54.9 & 53.0 & 58.4  \\ 
   ~ & ~ & ~ & \textcolor{greenberry}{(-5.5} & \textcolor{greenberry}{(-0.1)} & \textcolor{greenberry}{(-2.0)} & \textcolor{red}{(+3.4)}  \\  \hline
 \multirow{8}*{LLaMA2-13B} 
   & \multirow{2}*{Symbol} & \multirow{2}*{59.2} & 54.8 & 64.6 & 56.3 & 61.5   \\ 
   ~ & ~ & ~ & \textcolor{greenberry}{(-4.4)} & \textcolor{red}{(+5.4)} & \textcolor{greenberry}{(-2.9)} & \textcolor{red}{(+2.3)}  \\
   & \multirow{2}*{SCB} & \multirow{2}*{58.6} & 50.2 & 56.6 & 56.7 & 68.7  \\ 
   ~ & ~ & ~ & \textcolor{greenberry}{(-8.4)} & \textcolor{greenberry}{(-2.0)} & \textcolor{greenberry}{(-1.9)} & \textcolor{red}{(+10.1)}  \\  
   & \multirow{2}*{RSCB} & \multirow{2}*{58.7} & 54.3 & 61.7 & 62.7 & 57.5  \\ 
   ~ & ~ & ~ & \textcolor{greenberry}{(-4.4)} & \textcolor{red}{(+3.0)} & \textcolor{red}{(+4.0)} & \textcolor{greenberry}{(-1.2)}  \\  
  & \multirow{2}*{PIF} & \multirow{2}*{60.0} & 57.4 & 58.7 & 64.8 & 56.8  \\ 
   ~ & ~ & ~ & \textcolor{greenberry}{(-2.6)} & \textcolor{greenberry}{(-1.3)} & \textcolor{red}{(+4.8)} & \textcolor{greenberry}{(-3.2)}  \\  
\bottomrule
\end{tabular}
}
\end{center}
\caption{\label{table:MMLU results}The accuracy results after answer-moving attack on the LLMs with MMLU benchmarks.}
\end{table}
\setlength{\tabcolsep}{1.4pt}

\setlength{\tabcolsep}{4pt}
\begin{table}[t]
\begin{center}
\scalebox{0.65}{
\begin{tabular}{lccccccc}
\toprule
\multicolumn{2}{c}{\textbf{Move Golden to}}  & \textbf{orig} & \textbf{A} & \textbf{B} & \textbf{C} & \textbf{D} & \textbf{E} \\ 
\noalign{\smallskip} \hline

 \multirow{8}*{LLaMA2-7B} 
   & \multirow{2}*{Symbol} & \multirow{2}*{79.8} & 82.6 & 77.2 & 80.7 & 79.1 & 79.6  \\ 
   ~ & ~ & ~ & \textcolor{red}{(+2.8)} & \textcolor{greenberry}{(-2.6)} & \textcolor{red}{(+0.9)} & \textcolor{greenberry}{(-0.7)} &  \textcolor{greenberry}{(-0.2)} \\
   & \multirow{2}*{SCB} & \multirow{2}*{78.4} & 76.2 & 77.1 & 80.3 & 79.2 & 82.3  \\ 
   ~ & ~ & ~ & \textcolor{greenberry}{(-2.2)} & \textcolor{greenberry}{(-1.3)} & \textcolor{red}{(+1.9)} & \textcolor{red}{(+0.8)} & \textcolor{red}{(+3.9)}  \\ 
   & \multirow{2}*{RSCB} & \multirow{2}*{79.7} & 82.8 & 81.8 & 79.9 & 74.9 & 77.8  \\ 
   ~ & ~ & ~ & \textcolor{red}{(+3.1)} & \textcolor{red}{(+2.1)} & \textcolor{red}{(+0.2)} & \textcolor{greenberry}{(-4.8)} & \textcolor{greenberry}{(-1.9)}\\  
   & \multirow{2}*{PIF} & \multirow{2}*{79.4} & 79.8 & 77.6 & 81.7 & 77.3 & 77.8 \\ 
   ~ & ~ & ~ & \textcolor{red}{(+0.4)} & \textcolor{greenberry}{(-1.8)} & \textcolor{red}{(+2.3)} & \textcolor{greenberry}{(-2.1)} & \textcolor{greenberry}{(-1.6)} \\  \hline
 \multirow{8}*{LLaMA2-13B} 
   & \multirow{2}*{Symbol} & \multirow{2}*{81.5} & 81.3 & 76.2 & 86.2 & 82.9 & 83.7   \\ 
   ~ & ~ & ~ & \textcolor{greenberry}{(-0.2)} & \textcolor{greenberry}{(-5.3)} & \textcolor{red}{(+4.7)} & \textcolor{red}{(+1.4)} & \textcolor{red}{(+2.2)}  \\
   & \multirow{2}*{SCB} & \multirow{2}*{81.0} & 79.0 & 81.8 & 82.5 & 87.6  & 79.1 \\ 
   ~ & ~ & ~ & \textcolor{greenberry}{(-2.0)} & \textcolor{red}{(+0.8)} & \textcolor{red}{(+1.5)} & \textcolor{red}{(+6.6)} & \textcolor{greenberry}{(-1.9)}  \\  
   & \multirow{2}*{RSCB} & \multirow{2}*{80.9} & 81.8 & 82.9 & 82.5 & 77.9 & 78.8  \\ 
   ~ & ~ & ~ & \textcolor{red}{(+0.9)} & \textcolor{red}{(+2.0)} & \textcolor{red}{(+1.6)} & \textcolor{greenberry}{(-3.0)} & \textcolor{greenberry}{(-2.1)}  \\  
  & \multirow{2}*{PIF} & \multirow{2}*{82.2} & 79.9 & 82.0 & 82.5 & 83.4 &81.7  \\ 
   ~ & ~ & ~ & \textcolor{greenberry}{(-2.3)} & \textcolor{greenberry}{(-0.2)} & \textcolor{red}{(+0.3)} & \textcolor{red}{(+1.2)} & \textcolor{greenberry}{(-0.5)}  \\  
\bottomrule
\end{tabular}
}
\end{center}
\caption{\label{table:CSQA results}The accuracy results after answer-moving attack on the LLMs with CSQA benchmarks.}
\end{table}
\setlength{\tabcolsep}{1.4pt}

\section{Proof of The Relationship Between MCSB Capability and Selection Bias}
\label{sec:relationship proof}
For a single instance, we denote its PPA as $v_\text{ppa}$, its selection bias as $v_\text{bias}$, its accuracy as $v_\text{Acc}$. From Equation~\ref{eq:mubias},  $v_\text{bias}  = \frac{\sum_{i=1}^{K} (|v_{\text{Acc}_i} - v_{\text{Acc}_0}|)}{K}$. Considering that $v_{\text{Acc}_0}$ can only be 0 or 1 for a single instance, then $v_\text{bias}$ is:
\begin{equation}
\begin{aligned}
\label{eq:v_bias}
v_\text{bias}  =  
\left\{
\begin{array}{rcl}
\frac{\sum_{i=1}^{K} v_{\text{Acc}_i}}{K}       &      & v_{\text{Acc}_0 = 0}\\
1-\frac{\sum_{i=1}^{K} v_{\text{Acc}_i}}{K}     &      & v_{\text{Acc}_0 = 1}

\end{array} 
\right.
\end{aligned}
\end{equation}

On the other hand, $v_\text{ppa}=\frac{\max\limits_{k\sim K}(\sum\limits_{j=1}\limits^{K!}y_j=o_k)}{K!}$. Given a question with $K$ options, there are $K!$ distinct arrangements of these options, we now use a subset, always moving the golden option content to a specific symbol, to represent the overall distribution, then $v_\text{ppa}=\frac{\max\limits_{k\sim K}(\sum\limits_{j=1}\limits^{K}y_j=o_k)}{K}$. If the most frequently predicted option is the correct answer, $v_{\text{Acc}_0}$ will have a probability of $v_\text{ppa}$ to take the value of 1, and $v_\text{ppa}=\frac{\sum_{i=1}^{K} v_{\text{Acc}_i}}{K}$; else, 
$v_\text{ppa}=\frac{\sum_{i=1}^{K} (1-v_{\text{Acc}_i})}{K}$. Thus we define:
\begin{equation}
\begin{aligned}
\label{eq:v_ppa}
v_\text{ppa}  =  
\left\{\begin{array}{rcl}
\frac{\sum_{i=1}^{K} v_{\text{Acc}_i}}{K}       &      & v_{\text{Acc}_0 = 1}\\
1-\frac{\sum_{i=1}^{K} v_{\text{Acc}_i}}{K}     &      & v_{\text{Acc}_0 = 0}
\end{array} \right.
\end{aligned}
\end{equation}

From Equation~\ref{eq:v_bias},~\ref{eq:v_ppa}, we can observe that $v_\text{bias}$ and $v_\text{ppa}$ are inversely related with a probability of $v_\text{ppa}$. Generally, after SFT, LLMs tend to have higher $v_\text{ppa}$, Therefore providing theoretical support for our work.

\setlength{\tabcolsep}{4pt}
\begin{table*}
\small
\begin{center}
\scalebox{0.8}{
\begin{tabular}{p{0.15\linewidth}p{0.85\linewidth}}
\toprule
\multirow{3}*{\textbf{Question}}
&For which of these two scenarios does the main character (who uses I/me/my) do something clearly morally\\
~&wrong, according to ordinary moral standards in the US as of 2020? Scenario 1 | I slammed on my breaks to \\
~&miss the ball as it came in to the road. Scenario 2 | I taught my children to play the xylophone. 
\\ 
\noalign{\smallskip} \hline\noalign{\smallskip}
\textbf{Options}
&A:Wrong, Wrong B:Wrong, Not wrong C:Not wrong, Wrong D:Not wrong, Not wrong 
\\ 
\noalign{\smallskip} \hline\noalign{\smallskip}
\multirow{4}*{\textbf{Option Symbols}}
  &  $s_{1}$: A \\
  ~& $s_{2}$: B \\
  ~& $s_{3}$: C \\
  ~& $s_{4}$: D \\ 
\noalign{\smallskip} \hline\noalign{\smallskip}
\multirow{4}*{\textbf{Option contents}}
  & $o_{1}$: {Wrong, Wrong} \\ 
    ~& $o_{2}$: Wrong, Not wrong \\ 
    ~& $o_{3}$: Not wrong, Wrong \\
    ~& $o_{4}$: Not wrong, Not wrong \\ 
    \noalign{\smallskip} \hline \noalign{\smallskip}
\multirow{24}*{\textbf{Arrangements}}
  & $a_{1}$: A: Wrong, Wrong B: Wrong, Not wrong C: Not wrong, Wrong D: Not wrong, Not wrong \\
  ~& $a_{2}$: A: Wrong, Wrong B: Wrong, Not wrong C: Not wrong, Not wrong D: Not wrong, Wrong \\
  ~& $a_{3}$: A: Wrong, Wrong B: Not wrong, Wrong C: Wrong, Not wrong D: Not wrong, Not wrong \\
  ~& $a_{4}$: A: Wrong, Wrong B: Not wrong, Wrong C: Not wrong, Not wrong D: Wrong, Not wrong \\
  ~& $a_{5}$: A: Wrong, Wrong B: Not wrong, Not wrong C: Wrong, Not wrong D: Not wrong, Wrong \\
  ~& $a_{6}$: A: Wrong, Wrong B: Not wrong, Not wrong C: Not wrong, Wrong D: Wrong, Not wrong \\
  ~& $a_{7}$: A: Wrong, Not wrong B: Wrong, Wrong C: Not wrong, Wrong D: Not wrong, Not wrong \\
  ~& $a_{8}$: A: Wrong, Not wrong B: Wrong, Wrong C: Not wrong, Not wrong D: Not wrong, Wrong \\
  ~& $a_{9}$: A: Wrong, Not wrong B: Not wrong, Wrong C: Wrong, Wrong D: Not wrong, Not wrong \\
  ~& $a_{10}$: A: Wrong, Not wrong B: Not wrong, Wrong C: Not wrong, Not wrong D: Wrong, Wrong \\
  ~& $a_{11}$: A: Wrong, Not wrong B: Not wrong, Not wrong C: Wrong, Wrong D: Not wrong, Wrong \\
  ~& $a_{12}$: A: Wrong, Not wrong B: Not wrong, Not wrong C: Not wrong, Wrong D: Wrong, Wrong \\
  ~& $a_{13}$: A: Not wrong, Wrong B: Wrong, Wrong C: Wrong, Not wrong D: Not wrong, Not wrong \\
  ~& $a_{14}$: A: Not wrong, Wrong B: Wrong, Wrong C: Not wrong, Not wrong D: Wrong, Not wrong \\
  ~& $a_{15}$: A: Not wrong, Wrong B: Wrong, Not wrong C: Wrong, Wrong D: Not wrong, Not wrong \\
  ~& $a_{16}$: A: Not wrong, Wrong B: Wrong, Not wrong C: Not wrong, Not wrong D: Wrong, Wrong \\
  ~& $a_{17}$: A: Not wrong, Wrong B: Not wrong, Not wrong C: Wrong, Wrong D: Wrong, Not wrong \\
  ~& $a_{18}$: A: Not wrong, Wrong B: Not wrong, Not wrong C: Wrong, Not wrong D: Wrong, Wrong \\
  ~& $a_{19}$: A: Not wrong, Not wrong B: Wrong, Wrong C: Wrong, Not wrong D: Not wrong, Wrong \\
  ~& $a_{20}$: A: Not wrong, Not wrong B: Wrong, Wrong C: Not wrong, Wrong D: Wrong, Not wrong \\
  ~& $a_{21}$: A: Not wrong, Not wrong B: Wrong, Not wrong C: Wrong, Wrong D: Not wrong, Wrong \\
  ~& $a_{22}$: A: Not wrong, Not wrong B: Wrong, Not wrong C: Not wrong, Wrong D: Wrong, Wrong \\
  ~& $a_{23}$: A: Not wrong, Not wrong B: Not wrong, Wrong C: Wrong, Wrong D: Wrong, Not wrong \\
  ~& $a_{24}$: A: Not wrong, Not wrong B: Not wrong, Wrong C: Wrong, Not wrong D: Wrong, Wrong \\ 
  \noalign{\smallskip} \hline \noalign{\smallskip}
  
\multirow{24}*{\textbf{Model outputs}}
& $y_{1}$: \textcolor{red}{Wrong, Not wrong} \\
~& $y_{2}$: \textcolor{red}{Wrong, Not wrong} \\
~& $y_{3}$: Not wrong, Wrong \\
~& $y_{4}$: Not wrong, Wrong \\
~& $y_{5}$: \textcolor{red}{Wrong, Not wrong} \\
~& $y_{6}$: \textcolor{red}{Wrong, Not wrong} \\
~& $y_{7}$: Wrong, Wrong \\
~& $y_{8}$: Wrong, Wrong \\
~& $y_{9}$: Not wrong, Wrong \\
~& $y_{10}$: Wrong, Wrong \\
~& $y_{11}$: Not wrong, Wrong \\
~& $y_{12}$: Wrong, Wrong \\
~& $y_{13}$: Wrong, Wrong \\
~& $y_{14}$: \textcolor{red}{Wrong, Not wrong} \\
~& $y_{15}$: \textcolor{red}{Wrong, Not wrong} \\
~& $y_{16}$: \textcolor{red}{Wrong, Not wrong} \\
~& $y_{17}$: Wrong, Wrong \\
~& $y_{18}$: Wrong, Wrong \\
~& $y_{19}$: \textcolor{red}{Wrong, Not wrong} \\
~& $y_{20}$: Wrong, Wrong \\
~& $y_{21}$: \textcolor{red}{Wrong, Not wrong} \\
~& $y_{22}$: \textcolor{red}{Wrong, Not wrong} \\
~& $y_{23}$: Not wrong, Wrong \\
~& $y_{24}$: \textcolor{red}{Wrong, Not wrong} \\
 \noalign{\smallskip} \hline \noalign{\smallskip}
 
\multirow{4}*{\textbf{Frequency}}
& $o_{1}$(Wrong, Wrong): $8 \div 4! \approx 0.333$ \\ 
    ~& $o_{2}$(Wrong, Not wrong): \textcolor{red}{$11 \div 4! \approx 0.458$} \\ 
    ~& $o_{3}$(Not wrong, Wrong): $5 \div 4! \approx 0.208$ \\
    ~& $o_{4}$(Not wrong, Not wrong): $0 \div 4! = 0$ \\ 
\noalign{\smallskip} \hline \noalign{\smallskip}
\textbf{PPA result} & \textcolor{red}{$0.458$}\\ 
\bottomrule

\end{tabular}
}
\end{center}
\caption{\label{table:ppa cal} An example of the PPA calculation process for a single instance. Arrangements are the results of the arrangement of option contents. Model output $y_{i}$ refers to the answer produced by the model after the question and $a_{i}$ are input into it. The PPA metric does not consider whether the model performs tasks rightly. It measures the consistency of the model during the execution of tasks. }
\end{table*}
\setlength{\tabcolsep}{1.4pt}

\label{sec:permutation}
\end{document}